\documentclass[10pt]{article} 
\usepackage[preprint]{tmlr}

\usepackage{hyperref}
\usepackage{url}

\usepackage{amsmath}
\usepackage{amssymb}
\usepackage{multirow}
\usepackage{catchfile}
\usepackage{varwidth}
\usepackage{cleveref}
\usepackage[normalem]{ulem}
\usepackage{lipsum}
\usepackage{graphicx}
\usepackage{subcaption}
\usepackage{pifont}
\usepackage[font=small,labelfont=bf]{caption}
\usepackage{float}
\usepackage{booktabs}
\usepackage{subcaption}
\usepackage{listings}
\usepackage{natbib}
\usepackage{doi}
\usepackage{appendix}
\usepackage[dvipsnames]{xcolor}
\usepackage{bm}
\usepackage{amsthm}

\newtheorem{theorem}{Theorem}
\newtheorem{fact}[theorem]{Fact}

\makeatletter
\newcommand*\bigcdot{\mathpalette\bigcdot@{.5}}
\newcommand*\bigcdot@[2]{\mathbin{\vcenter{\hbox{\scalebox{#2}{$\m@th#1\bullet$}}}}}

\newcommand{\ei}{e_i}
\newcommand{\eistar}{e_i^\star}
\newcommand{\eithree}{e_i^{(3)}}
\newcommand{\h}{h}
\newcommand{\hthree}{h^{(3)}}

\newcommand{\meanembedding}{\mu}
\newcommand{\meanlogit}{\overline{l}}
\newcommand{\stdlogit}{\sigma_l}
\newcommand{\logitcentered}{l^\star}

\newcommand{\meanlogitcentered}{\overline{l^\star}}
\newcommand{\stdlogitcentered}{\sigma_{l^\star}}
\newcommand{\ratioembeddingi}{r_{\rm emb}^{(i)}}
\newcommand{\ratiocosine}{r_{\rm cos}}
\newcommand{\ratiologit}{r_{\rm log}}

\usepackage[rightcaption]{sidecap}
\definecolor{clrbaseline}{HTML}{FF2C00}
\definecolor{clrweightdecay}{HTML}{845B97}
\definecolor{clrzloss}{HTML}{0C5DA5}
\definecolor{clrsoftcapping}{HTML}{9e9e9e}
\definecolor{clrmuloss}{HTML}{FF9500}
\definecolor{clrmucentering}{HTML}{00B945}

\title{Output Embedding Centering for Stable LLM Pretraining}

\author{\name Felix Stollenwerk \email felix.stollenwerk@ai.se \\
      \addr AI Sweden
      \AND
      \name Anna Lokrantz \email anna.lokrantz@ai.se \\
      \addr AI Sweden
      \AND
      \name Niclas Hertzberg \email niclas.hertzberg@ai.se\\
      \addr AI Sweden
      }

\def\month{MM}  
\def\year{YYYY} 
\def\openreview{\url{https://openreview.net/forum?id=XXXX}} 

\begin{document}

\maketitle

\begin{abstract}
Pretraining of large language models is not only expensive but also prone to certain training instabilities. A specific instability that often occurs at the end of training is output logit divergence. The most widely used mitigation strategies, z-loss and logit soft-capping, merely address the symptoms rather than the underlying cause of the problem. In this paper, we analyze the instability from the perspective of the output embeddings' geometry and identify anisotropic embeddings as its source. Based on this, we propose \textit{output embedding centering (OEC)} as a new mitigation strategy, and demonstrate that it suppresses output logit divergence. OEC can be implemented in two different ways: as a deterministic operation called $\mu$-centering, or a regularization method called $\mu$-loss. Our experiments show that both variants outperform z-loss in terms of training stability, while being on par with logit soft-capping. This holds true both in the presence and the absence of weight tying. As a secondary result, we find that $\mu$-loss is significantly less sensitive to regularization hyperparameter tuning than z-loss.
\end{abstract}

\section{Introduction}
\label{sec:introduction}

Large language models (LLMs) have shown great capabilities in solving many different types of tasks. However, instability during the most computationally expensive phase of pretraining LLMs is a recurring issue \citep{chowdhery2022palmscalinglanguagemodeling, takase2025spike, pmlr-v202-dehghani23a}, often resulting in a significant amount of wasted compute. There are several types of training instabilities, e.g. extremely large attention logits \citep{pmlr-v202-dehghani23a} or divergence of the output logits in the language modeling head \citep{chowdhery2022palmscalinglanguagemodeling}. \citet{wortsman2023smallscaleproxieslargescaletransformer} have shown that these training instabilities can be reproduced in a proxy setting of small-scale models in combination with high learning rates. This allows to study the effect of mitigation strategies in isolation on a small scale, and draw conclusions for large-scale pretraining. 
In this work, we adopt their approach and specifically address the problem of output logit divergence.

\paragraph{Language Modeling Head}
We consider decoder-only Transformer models \citep{vaswani2017attention, radford2018improving}, in which the language modeling head is the final component responsible for mapping the final hidden state to a probability distribution over the tokens in the vocabulary. 
Following the notation of \citet{stollenwerk-stollenwerk-2025-better}, the standard language modeling head is defined by the following equations:
\begin{align}
\mathcal{L} = - \log{(p_t)} 
\qquad \qquad
p_t = \frac{\exp{(l_t)}}{\sum_{j=1}^V \exp{(l_j)}} 
\qquad \qquad
l_i = \ei \bigcdot \h 
\label{eq:lmhead}
\end{align}
$\mathcal{L}\in \mathbb{R}_{\geq 0}$ is the loss for next token prediction, while $p_t \in [0,1]$ represents the probability assigned to the true token $t \in \mathcal{V}$. Here, $\mathcal{V}\equiv \{1, \ldots, V\}$, where $V$ is the size of the vocabulary. The logits and output embeddings for each token $i \in \mathcal{V}$ are denoted by $l_i \in \mathbb{R}$ and $\ei \in \mathbb{R}^H$, respectively, with $H$ being the dimension of the model's hidden space. The final hidden state is given by $\h \in \mathbb{R}^H$.
The output embeddings $\ei$ can either be learned independently or tied to the input embeddings \citep{press-wolf-2017-using}.

\paragraph{Output Logit Divergence}

The problem of output logit divergence can be traced back to numerical instabilities, caused by roundoff errors that are aggravated by exponential functions as present in\footnote{We refer to subequations in a single line using lowercase letters that follow the order from left to right.} Eq.~\ref{eq:lmhead}b. Following the discussion and example given in \cite{zoph2022stmoedesigningstabletransferable}, consider the logits $l = (128.5, 128, \ldots, 128) \in \mathbb{R}^{11}$. For the sake of numerical stability, each logit is subtracted by the largest logit $l_1$ in the softmax, resulting in $p_1 = \frac{\exp(0)}{\exp(0) + 10 \cdot \exp(-0.5)} \approx 0.142$. If $l_0$ is subject to a roundoff error and incorrectly assigned the value $128$, one gets $p_1 = \frac{\exp(0)}{11 \cdot \exp(0)} \approx 0.091$ instead. This large numerical mismatch can lead to training instabilities. As roundoff errors are more likely to occur for large numbers (see e.g. \cite{zoph2022stmoedesigningstabletransferable} for a detailed discussion), these instabilities can be suppressed by decreasing the size of the absolute logits $\left| l_i \right|$. This is the principle that all mitigation strategies\footnote{Increasing the floating point precision is not considered here due to its higher computational demands.} are based on.

\paragraph{Mitigation Strategies}

A technique that is not specific to output logit divergence, but rather aims to suppress large weights in general, is weight decay. It can also be applied to the output embeddings $\ei$ in order to control the logits via Eq.~\ref{eq:lmhead}c. For each embedding weight $\theta = e_{ij}$, it leverages a weight decay term $-\lambda \theta$, in addition to the ordinary optimizer update $u_\theta$, in order to reduce the size of $\theta$:
\begin{align}
    \theta \leftarrow \left( 1 - \lambda \right) \theta - \eta~u_\theta
\end{align}
The above variant, which we will consider with
\begin{align}
    \lambda &= 10^{-4}
    \label{eq:lambda_wd}
\end{align}
as in \citet{wortsman2023smallscaleproxieslargescaletransformer}, was proposed by \citet{loshchilov2018decoupled} and features a weight decay term that is independent of the learning rate $\eta$.
The most widely adopted solution targeted specifically at the problem of divergent output logits is \textit{z-loss}, introduced by \citet{chowdhery2022palmscalinglanguagemodeling}. Denoting the denominator of Eq.~\ref{eq:lmhead}b by
\begin{align}
Z := \sum_{j=1}^V \exp{(l_j)}
\label{eq:Z}
\end{align} 
z-loss adds a regularization term of the form
\begin{align}
\mathcal{L}_z := \lambda \cdot \log^2 \left( Z \right)
\qquad \text{with} \qquad
\lambda = 10^{-4}
\label{eq:zloss}
\end{align}
It has proven an effective measure to stabilize the training process, also in the proxy setting of \citet{wortsman2023smallscaleproxieslargescaletransformer}.
Consequently, z-loss has been utilized in several recent models \citep{olmo20252olmo2furious, chameleonteam2025chameleonmixedmodalearlyfusionfoundation, pmlr-v162-wang22u, OLMo3_2025}. Similarly, Baichuan 2 \citep{yang2025baichuan2openlargescale} introduced a variant of z-loss, max-z loss, that penalizes the square of the maximum logit value. 
In contrast to adding auxiliary losses, Gemma 2 \citep{gemmateam2024gemma2improvingopen} enforces bounds via \textit{logit soft-capping}
\begin{align}
   l_i \leftarrow c \cdot \tanh \left( l_i / c \right)
   \label{eq:logit_softcapping}
\end{align}
which is applied between Eq.~\ref{eq:lmhead}b and Eq.~\ref{eq:lmhead}c. This model intervention confines the logits to a fixed numerical range determined by the hyperparameter $c$ with default value $c=30$.
Another method, NormSoftMax \citep{10189242}, proposes a dynamic temperature scaling in the softmax function based on the distribution of the logits.
The above methods all have in common that they address the symptoms rather than the cause of large output logits. 
In an attempt to identify the latter, we will examine the role of the output embeddings\footnote{The final hidden states are arguably less relevant in this context, as they are usually normalized.}, which affect the output logits via Eq.~\ref{eq:lmhead}c.

\paragraph{Anisotropic Embeddings}

A well-known phenomenon exhibited by the embeddings of Transformer models is that they typically do not distribute evenly across the different dimensions in hidden space. This problem of anisotropy\footnote{Note that the discussion here is based on the definition of isotropy from \citet{arora-etal-2016-latent} and \citet{mu2018allbutthetopsimpleeffectivepostprocessing}, which is mean-sensitive.} was first described by \citet{gao2019representationdegenerationproblemtraining}. At the time, the understanding was that the embeddings occupy a narrow cone in hidden space. Several regularization methods have been proposed to mitigate the problem, e.g. cosine regularization \citep{gao2019representationdegenerationproblemtraining}, Laplace regularization \citep{zhang-etal-2020-revisiting} and spectrum control \citep{Wang2020ImprovingNL}. 
\citet{bis2021tmic} showed that embeddings are actually near-isotropic around their center, and argued that the observed anisotropy is mainly due to a common shift of the embeddings away from the origin. 
Recently, \citet{stollenwerk-stollenwerk-2025-better} identified the root cause of this phenomenon; they showed that it is the second moment in Adam that causes the common shift of the embeddings. Furthermore, their analysis reveals that the phenomenon stems from the output embeddings rather than the input embeddings, in accordance with the observations reported in \citet{machina-mercer-2024-anisotropy}.

\paragraph{Our Contributions}

\begin{itemize}
\item \textit{Analysis:} We analyze the role of anisotropic embeddings in causing output logit divergence.
\item \textit{Methods:} We propose \textit{output embedding centering (OEC)}, a theoretically well-motivated mitigation strategy with two implementations, \textit{$\mu$-centering} and \textit{$\mu$-loss}, that both keep the output embeddings and logits centered around zero.
\item \textit{Training Stability:} We provide a comprehensive and systematic comparison of OEC and existing mitigation strategies in the small-scale proxy setting of \citet{wortsman2023smallscaleproxieslargescaletransformer}. It is shown that our methods lead to an equal or better learning rate sensitivity, which is expected to transfer to (more) effective stabilization of large-scale LLM pretraining.
\item \textit{Hyperparameter Sensitivity:} We show that our regularization method $\mu$-loss is significantly less sensitive to the regularization hyperparameter than z-loss.
\end{itemize}
 
\section{Output Embedding Centering}
\label{sec:theory}

\subsection{Output Embeddings and Logits}

In this section, we examine the relationship between the output embeddings $\ei$ and logits $l_i$, see Eq.~\ref{eq:lmhead}c. 
We start with the mean word embedding $\meanembedding$ and the mean logit $\meanlogit$:
\begin{align}
\meanembedding &= \frac{1}{V} \sum_{i=1}^V \ei 
\qquad \qquad
\meanlogit
= \frac{1}{V} \sum_{i=1}^V l_i
\label{eq:mu_meanlogit}
\end{align}
Their relationship is expressed by the following elementary identity, which follows directly from the linearity of the dot product.
\begin{fact}
\label{lemma_mean_logit}
The mean logit is proportional to the mean embedding:
\begin{align}
\meanlogit &= \meanembedding \bigcdot \h
\label{eq:mean_logit_expression}
\end{align}  
\end{fact}

Hence, controlling the output embeddings provides a means to control the logits. This insight lays the foundation for \textit{output embedding centering} (OEC). The idea behind OEC is to ensure that the mean output embedding $\meanembedding$ (cf.~Eq.~\ref{eq:mu_meanlogit}a) is bound to the origin, suppressing the common shift of the embeddings (cf.~Sec.~\ref{sec:introduction}) and logit growth. 
OEC comes in two variants, \textit{$\mu$-centering} and \textit{$\mu$-loss}, which we will introduce next.

\subsection{$\mu$-centering}
\label{sec:theory_mucentering}

OEC can be implemented in a deterministic, hyperparameter-free manner by subtracting the mean output embedding $\meanembedding$ from each output embedding $\ei$, creating new output embeddings $\eistar$ after each optimization step:
\begin{align}
\eistar &= \ei - \meanembedding
\label{eq:output_embedding_centering}
\end{align}
This variant, called \textit{$\mu$-centering}, has some simple, well-known implications that can be summarized as follows:
\begin{fact}
\label{lemma_mec}
Let $l$ and $\logitcentered$ denote the output logits before and after $\mu$-centering, respectively.
\begin{enumerate}
\item[(i)] The mean output logit after $\mu$-centering is zero: $\meanlogitcentered = 0$
\item[(ii)] The output logits standard deviation is invariant under $\mu$-centering: $\stdlogitcentered = \stdlogit$
\item[(iii)] The output probabilities, loss and gradients are invariant under $\mu$-centering. This implies that no model parameters other than the output embeddings are affected during training. 
\end{enumerate}
\end{fact}

\begin{SCfigure}[1][tb]
\centering
\includegraphics[scale=0.6, trim={18mm 8mm 0 8mm}, clip]{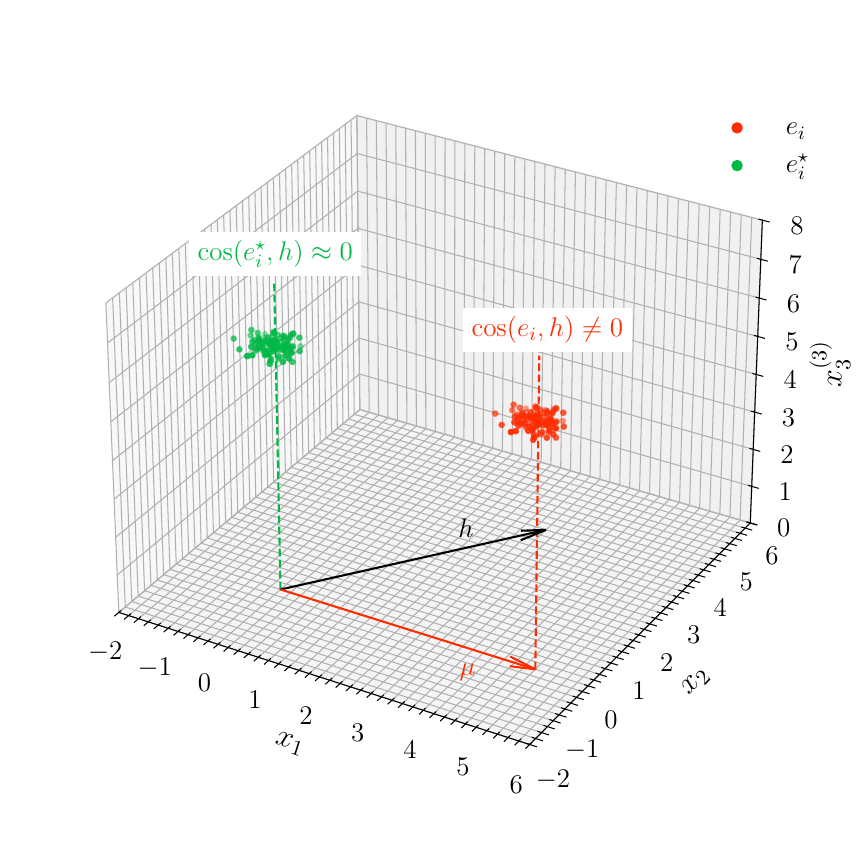}
\hfill
\caption{Illustration of the effect of $\mu$-centering on the logits. The vectors $\ei, \eistar, \meanembedding, \h \in \mathbb{R}^{H}$ are each shown in a norm-preserving projection, where an $H$-dimensional vector $x = (x_1, x_2, x_3, \ldots, x_H)$ is projected onto 3 dimensions, $x^{(3)} = (x_1, x_2, x_3^{(3)})$ with $x_3^{(3)} = \left(\sum_{i=3}^H x_i^2\right)^{1/2}$. This projection preserves angles, cosine similarities and dot products between two vectors if and only if one of them lies in the $(x_1, x_2)$-plane. As $x_1$ and $x_2$ are by definition chosen to be coordinates of ${\rm span}(\meanembedding, \h)$, the preservation holds in particular for the embeddings and the final hidden state, and thus the logits $l_i = \ei \bigcdot \h = \eithree \bigcdot \hthree$. The red points display the projection of vectors $\ei$ in $H = 1024$ dimensions that are near-isotropic around a shifted mean $\meanembedding \neq 0$ \citep{bis2021tmic}. An alignment of $\meanembedding$ and $\h$ leads to $ \cos \left( \ei, \h \right) \approx \cos \left( \meanembedding, \h \right) \neq 0$, while $\mu$-centering results in $\cos \left( \eistar, \h \right) \approx 0$. Together, this gives $\ratiocosine < 1$, which is the decisive factor leading to $\ratiologit < 1$ (see Fact~\ref{lemma_global_logit_bounds}).}
\label{fig:mu_centering_alignment}
\end{SCfigure}

In addition, for strongly anisotropic output embeddings, $\mu$-centering reduces the global logits bounds:
\begin{align}
    \max_i \left| l_{i}^\star \right| 
    < 
    \max_i \left| l_{i} \right|
    \label{eq:global_logit_bounds_informal}
\end{align}
The mechanism that leads to Eq.~\ref{eq:global_logit_bounds_informal} is expressed by Fact~\ref{lemma_global_logit_bounds}.
\begin{fact}
\label{lemma_global_logit_bounds}
Define the ratios
\begin{align}
\ratioembeddingi 
:=
\frac{\| \eistar \|}{\| \ei \|}     
\qquad \qquad
\ratiocosine 
:= \frac{\max_i \left| \cos \left( \h, \eistar \right) \right|}{\max_i \left| \cos \left( \h, \ei \right) \right|} 
\qquad \qquad
\ratiologit := \frac{\max_i \left| l_{i}^\star \right|}{\max_i \left| l_{i} \right|}
\label{eq:def_r}
\end{align}
If 
\begin{align}
    \ratioembeddingi
    < 1 \quad \forall~i \in \mathcal{V}
    \quad \qquad {\rm and} \quad \qquad
    \ratiocosine 
    < 1
    \label{eq:empirical}
\end{align}
then
\begin{align}
    \ratiologit < 1
\label{eq:global_logit_bounds}
\end{align}  
\end{fact}
\begin{proof} 
\begin{align}
\ratiologit 
&\stackrel{(\ref{eq:def_r}c)}{=} \frac{\max_i \left| l_{i}^\star \right|}{\max_i \left| l_{i} \right|}
\stackrel{(\ref{eq:lmhead}c, \ref{eq:output_embedding_centering})}{=} \frac{\max_i \left| \eistar \bigcdot \h \right|}{\max_i \left| \ei \bigcdot \h \right|} 
= \frac{\max_i \left\{ \| \h \| \cdot \| \eistar \| \cdot \left| \cos \left( \h, \eistar \right) \right| \right\}}{\max_i \left\{ \| \h \| \cdot \| \ei \| \cdot \left| \cos \left( \h, \ei \right) \right| \right\}} \nonumber \\
&\stackrel{(\ref{eq:empirical}a)}{<} \frac{\| \h \|}{\| \h \|} \cdot \frac{\max_i \left\{ \| \ei \| \cdot \left| \cos \left( \h, \eistar \right) \right| \right\}}{\max_i \left\{ \| \ei \| \cdot \left| \cos \left( \h, \ei \right) \right| \right\}} 
= \frac{\max_i \left| \cos \left( \h, \eistar \right) \right|}{\max_i \left| \cos \left( \h, \ei \right) \right|}
\stackrel{(\ref{eq:def_r}b)}{=} \ratiocosine
\stackrel{(\ref{eq:empirical}b)}{<} 1 \nonumber
\end{align}  
\end{proof}
As we will see in Sec.~\ref{sec:results}, the conditions in Eq.~\ref{eq:empirical} are fulfilled for training setups that are prone to output logit divergence. Notably, the dominant effect leading to output logits suppression by $\mu$-centering stems from $\ratiocosine$, which we will find to be quite small, $\ratiocosine \ll 1$. This, in turn, can be traced back to an alignment of the final hidden state with the embeddings, $\cos \left( \meanembedding, \h \right) \neq 0$, as illustrated in Fig.~\ref{fig:mu_centering_alignment}.
In summary, $\mu$-centering centers the logits around zero (Fact~2) and limits their magnitude (Fact~3), thereby reducing the probability of rounding errors leading to training instability.

\subsection{$\mu$-loss}
\label{sec:theory_muloss}

Instead of $\mu$-centering, we can also enforce OEC approximately by adding a regularization \textit{$\mu$-loss} of the form
\begin{align}
\mathcal{L}_\mu &= \lambda \cdot \meanembedding \bigcdot \meanembedding 
\label{eq:mer}
\end{align}
Here, $\lambda \in \mathbb{R}^+$ is a hyperparameter that is set to 
\begin{align}
\lambda = 10^{-4}
\label{eq:lambda_default}
\end{align}
by default, as in the case of weight decay (see Eq.~\ref{eq:lambda_wd}) and z-loss (see Eq.~\ref{eq:zloss}).
While $\mu$-loss may offer more flexibility, $\mu$-centering has the advantage of being strikingly simple, hyperparameter-free, deterministic and free of side effects (see Fact~\ref{lemma_mec}).

\subsection{Comparison to Existing Mitigation Strategies}

We conclude this section by comparing the presented OEC methods, $\mu$-centering and $\mu$-loss, with the most widely used existing mitigation strategies: weight decay, z-loss and logit soft-capping (Sec.~\ref{sec:introduction}).
Logit soft-capping stands out as the only method that represents a model intervention, i.e. an adjustment of the model architecture. The other methods, in contrast, leave the model architecture unchanged and intervene in the training process instead.
z-loss, on the other hand, is exceptional in the sense that it does not suppress positive and negative output logit divergences equally. 
This can easily be seen by considering $\mathcal{L}_z$ as a function of a single logit $l_i$, with all other logits fixed:
\begin{align}
\mathcal{L}_z (l_i) \stackrel{(\ref{eq:Z}, \ref{eq:zloss})}{\propto}
\log^2 \left( \exp{(l_i)} + \sum_{j \neq i} \exp{(l_j)} \right)
\label{eq:zloss_function_1D}
\end{align}
Given that the sum over $j \neq i$ is positive, this function is clearly non-symmetric: 
\begin{align}
\mathcal{L}_z (l_i) \neq \mathcal{L}_z (-l_i)
\end{align}
In particular, z-loss does neither suppress single, negative logit divergences, $l_i \to -\infty$ nor enforce $\meanlogit = 0$. We refer the reader to App.~\ref{app:zloss} for additional details and illustrations.
The other methods, however, treat positive and negative divergences equally. Logit soft-capping, for instance, uses an odd function (see Eq.~\ref{eq:logit_softcapping}), while $\mu$-loss is quadratic in $\meanembedding$ (see Eq.~\ref{eq:mer}) and thus invariant under $\meanembedding \to -\meanembedding$ and $\meanlogit \to -\meanlogit$.
Tab.~\ref{tab:regularization_methods_effect} summarizes the means by which the discussed mitigation strategies prevent logit divergence.
\begin{table*}[hb]
\centering
\scriptsize
\caption{Overview of methods and means by which logit divergences are suppressed. Symmetry refers to whether positive and negative output logit divergences are treated equally.}
\begin{tabular}{lcccc}
     \toprule 
     name & intervention & means & symmetry & anisotropy suppression \\ \midrule
     logit soft-capping & model & element-wise transformation & yes & no \\
     weight decay & training & loss regularization & yes & no \\
     z-loss & training & loss regularization & no & no \\
     $\mu$-loss & training & loss regularization & yes & yes \\
     $\mu$-centering & training & parameter shift & yes & yes \\ \bottomrule
\end{tabular}
\label{tab:regularization_methods_effect}
\end{table*}
The general theoretical advantages of $\mu$-loss and $\mu$-centering are their simplicity and the fact that they have a theoretical foundation that addresses the root cause of the problem. In addition, they suppress positive and negative divergences equally (unlike z-loss), and pose a minimal interference only at training time, without changing the model itself (unlike logit soft-capping).

\section{Experiments}
\label{sec:experiments}

Our approach to studying training stability with regard to output logit divergence primarily follows the small-scale proxy setup from \citet{wortsman2023smallscaleproxieslargescaletransformer}. In particular, we train dense decoder models with a modern Transformer architecture \citep{vaswani2017attention} on 13.1 billion tokens for 100000 steps, using 7 different learning rates: 
\begin{align}
\eta \in \{ \text{3e-4, 1e-3, 3e-3, 1e-2, 3e-2, 1e-1, 3e-1} \}
\label{eq:eta}
\end{align}
However, there are also a number of differences. We use FineWeb \citep{penedo2024fineweb} and the GPT-2 tokenizer \citep{Radford2019LanguageMA} with a vocabulary size of $V = 50304$. Our 5 model sizes,
\begin{align}
N \in \{ 16{\rm M}, 29{\rm M}, 57{\rm M}, 109{\rm M}, 221{\rm M} \}
\label{eq:model_sizes}
\end{align}
and the corresponding specifications (e.g. widths, number of layers and attention heads) are taken from \citet{porian2025resolvingdiscrepanciescomputeoptimalscaling}. 
The experiments are conducted with and without weight tying:
\begin{align}
\text{weight tying} \in \{ \text{false}, \text{true} \}
\label{eq:weight_tying_combinations}
\end{align}
In addition, we use SwiGLU hidden activations \citep{shazeer2020glu} and a non-truncated Xavier weight initialization \citep{Glorot2010UnderstandingTD}.
Further details on model architecture and hyperparameters are provided in App.~\ref{app:hyperparameters}.
For each of the 7 x 5 x 2 = 70 combinations of learning rate, model size and weight tying defined by Eq.~\ref{eq:eta}-\ref{eq:weight_tying_combinations}, we train 6 different models: A baseline model with the standard language modeling head, and models using weight decay, z-loss, logit soft-capping, $\mu$-loss as well as $\mu$-centering. 
In order to compare the variants, we consider the dependency of the test loss on the learning rate, and evaluate the learning rate sensitivity as defined in \citet{wortsman2023smallscaleproxieslargescaletransformer}: 
    \begin{align}
        {\rm LRS} &= \mathbb{E}_{\eta} \left[ \min (\mathcal{L}(\eta) , \mathcal{L}_0) - \min_\eta \mathcal{L} \right]
        \label{eq:lr_sensitivity}
    \end{align}
Here, $\eta$ is the learning rate from Eq.~\ref{eq:eta} and $\mathcal{L}_0$ denotes the loss at initialization time.
The evaluation in terms of LRS is done independently for each combination of model size and weight tying, Eq.~\ref{eq:model_sizes} and \ref{eq:weight_tying_combinations}. The learning rate sensitivity in this small-scale setting has been shown to be a valuable proxy for training stability in the large-scale setting \citep{wortsman2023smallscaleproxieslargescaletransformer}.    
Additionally, we investigate the dependency of a few other quantities on the learning rate for the purpose of analyzing the functionality of the different methods, as well as to empirically verify Facts~\ref{lemma_mean_logit}, \ref{lemma_mec} and \ref{lemma_global_logit_bounds}. Firstly, we consider the norm $\| \meanembedding \|$ of the mean embedding (see Eq.~\ref{eq:mu_meanlogit}a).
Secondly, we sample $1.3 \cdot 10^5$ final hidden states $h$ and logit vectors $l$ using our models and the test data to compute the logit mean $\meanlogit$ (Fact~\ref{lemma_mean_logit}), the logits standard deviation $\stdlogit$ (Fact~\ref{lemma_mec}) and the ratios $\ratiocosine$ and $\ratiologit$ (Fact~\ref{lemma_global_logit_bounds}).
Finally, we determine $\ratioembeddingi$ for each token in the vocabulary (Fact~\ref{lemma_global_logit_bounds}).

\section{Results}
\label{sec:results}

\subsection{Training Stability}

\begin{figure*}[t]
\centering
\includegraphics[scale=0.56]{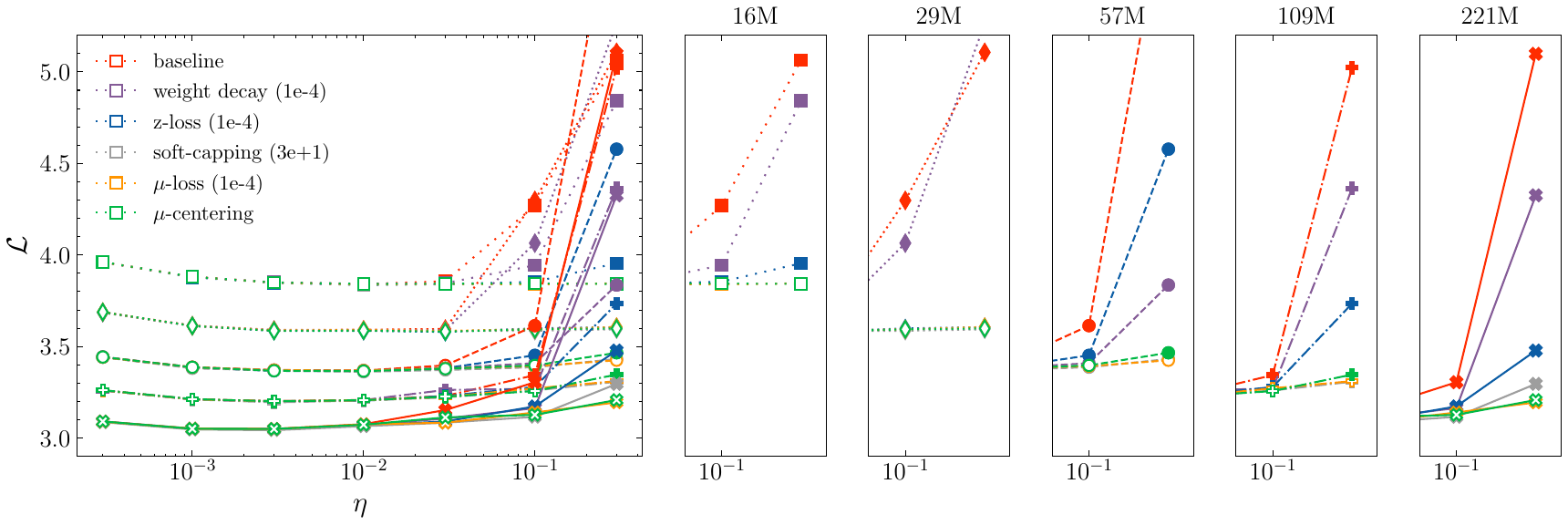}
\captionof{figure}{Dependency of the loss $\mathcal{L}$ on the learning rate $\eta$, for models without weight tying. The panel on the very left shows all results together, while the others focus on the largest two learning rates and a single model size $N$ (specified in the respective title). Filled and empty markers indicate divergence and convergence, respectively.}
\label{fig:wortsman}
\vspace{4mm}
\begin{minipage}[b]{.42\linewidth}
\centering
\includegraphics[scale=0.5]{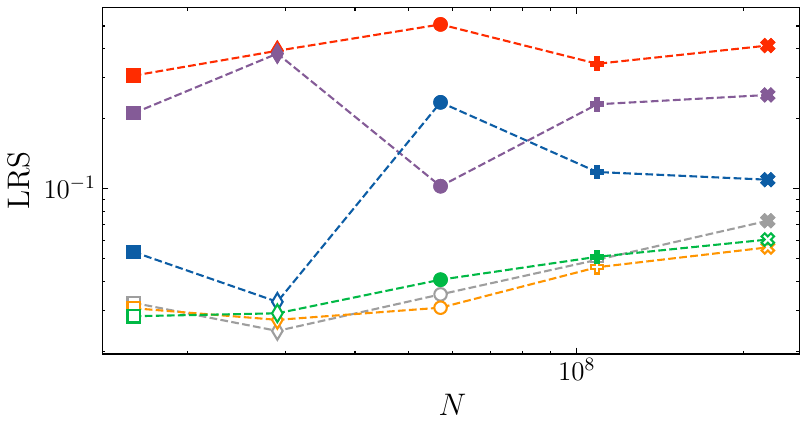}
\captionof{figure}{Learning rate sensitivity LRS for models without weight tying. Filled markers indicate that the training for at least one learning rate $\eta$ diverged. The colors represent the different variants as in Fig.~\ref{fig:wortsman}.}
\label{fig:wortsman_lrs}
\end{minipage}\hfill
\begin{minipage}[b]{.55\linewidth}
\centering
\scriptsize
\captionof{table}{Learning rate sensitivity LRS for models without weight tying. Underlined numbers indicate that the training for all learning rates converged, corresponding to the empty markers in Fig.~\ref{fig:wortsman_lrs}. In addition, the best result for each model size $N$ is highlighted in bold.}
\begin{tabular}{rcccccc}
    \multicolumn{7}{c}{LRS ($\downarrow$) ~|~ weight tying = false} \\
    \toprule
    $N$ & \colorbox{clrbaseline!30}{basel.} & \colorbox{clrweightdecay!30}{w.-d.} & \colorbox{clrzloss!30}{z-loss} & \colorbox{clrsoftcapping!30}{soft-c.} & \colorbox{clrmuloss!30}{$\mu$-loss} & \colorbox{clrmucentering!30}{$\mu$-cent.}  \\ \midrule
    16M & 0.306 & 0.211 & 0.054 & \underline{0.032} & \underline{0.031} & \underline{\textbf{0.028}} \\ 
    29M & 0.391 & 0.381 & \underline{0.033} & \underline{\textbf{0.024}} & \underline{0.027} & \underline{0.029} \\ 
    57M & 0.508 & 0.102 & 0.235 & \underline{0.035} & \underline{\textbf{0.031}} & 0.041 \\ 
    109M & 0.344 & 0.231 & 0.118 & \underline{0.050} & \underline{\textbf{0.046}} & 0.051 \\ 
    221M & 0.412 & 0.253 & 0.109 & 0.073 & \underline{\textbf{0.056}} & \underline{0.061} \\ \bottomrule
\end{tabular}
\vspace{4mm}
\label{tab:wortsman_lrs}
\end{minipage}
\end{figure*}

The results of our experiments without weight tying are shown in Fig ~\ref{fig:wortsman}.
As expected, the non-regularized baseline is the first to diverge with larger learning rates. Weight decay leads to weak improvements only. Interestingly, z-loss leads to occasional divergences as well, given a large enough learning rate\footnote{At first glance, this might seem to contradict the results from \citet{wortsman2023smallscaleproxieslargescaletransformer}. However, a thorough look at their Fig.~3 reveals a similar behavior for z-loss.}.
Meanwhile, the models using OEC or logit soft-capping are not affected by any significant training instabilities. This is also reflected in Tab.~\ref{tab:wortsman_lrs} and Fig.~\ref{fig:wortsman_lrs}, which show that these methods exhibit a lower learning rate sensitivity than z-loss, for all models sizes.
In our experiments, $\mu$-loss yields the best learning rate sensitivity. Its performance can even be further reduced by increasing the regularization parameter $\lambda$ from Eq.~\ref{eq:lambda_default}, see App.~\ref{app:ablations} for details. However, since the occurrence of output logit divergence is stochastic, minor differences between the different mitigation strategies should not be over-interpreted. Hence, we conclude conservatively that the ability of the different methods to suppress output logit divergence can be ranked as follows:
\begin{equation}
\boxed{
\mu\text{-centering} ~\approx~ \mu\text{-loss} ~\approx~ \text{logit soft-capping} ~>~ \text{z-loss} ~>~ \text{weight decay} ~>~ \text{baseline} 
}
\label{eq:main_conclusion}
\end{equation}
The results for models using weight tying are quite similar and shown in App.~\ref{app:wt}. 
In addition, we find that our methods' computational overhead is negligible, see App.~\ref{app:additional_results} for details.

\subsection{Analysis and Empirical Verification of Theory}

\begin{figure*}[ph]
\centering
\begin{minipage}[t]{.98\linewidth}
    \centering
    \includegraphics[scale=0.5]{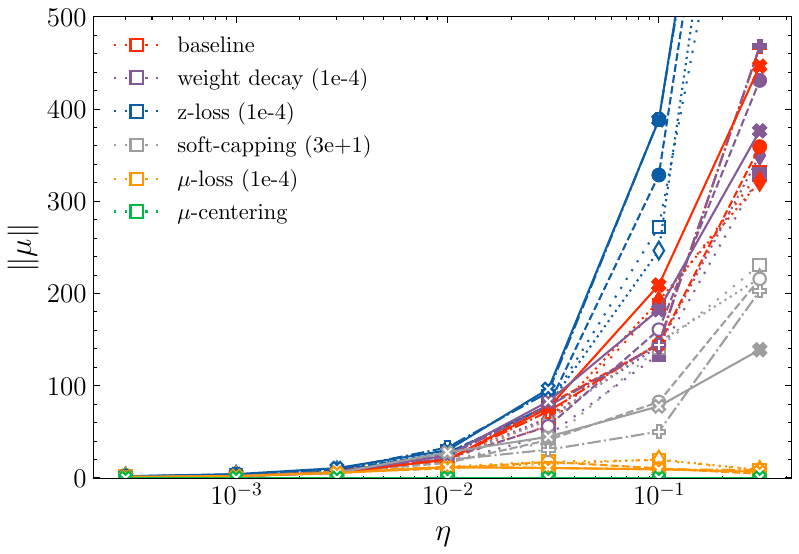}
    \hspace{4mm}
    \includegraphics[scale=0.5]{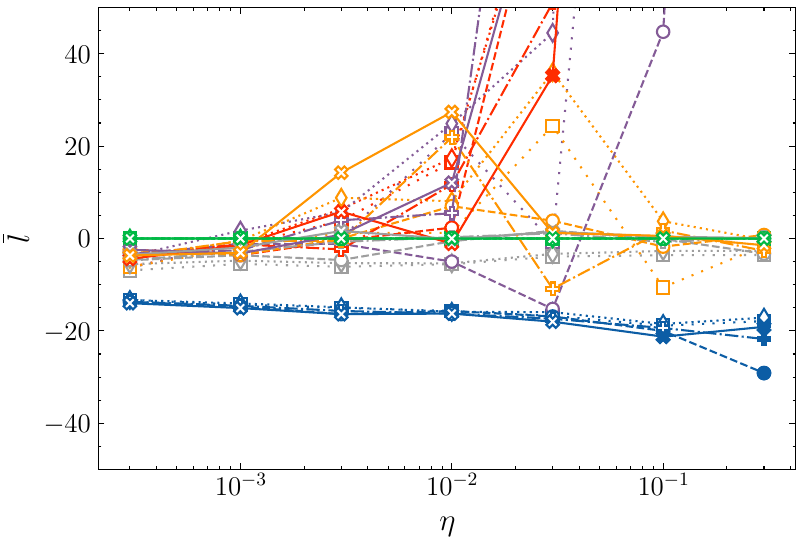} \\ 
    \vspace{2mm} 
    \hspace{0.5mm}
    \includegraphics[scale=0.5]{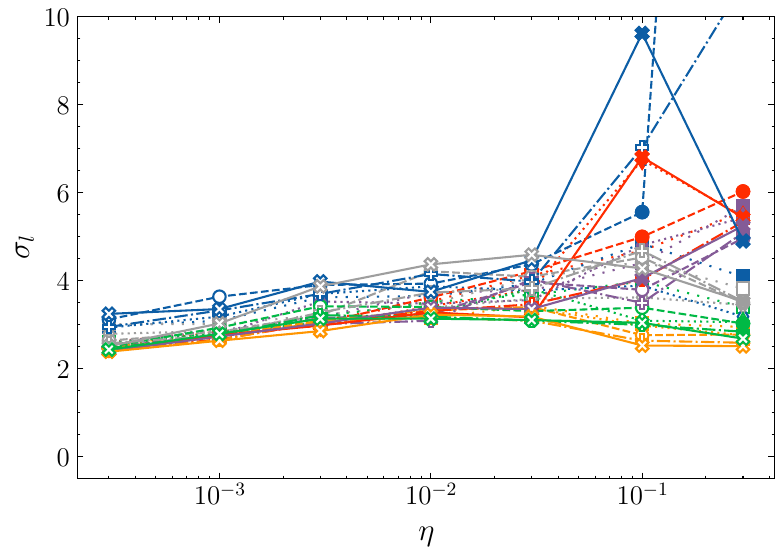} 
    \hspace{3mm}
    \includegraphics[scale=0.5]{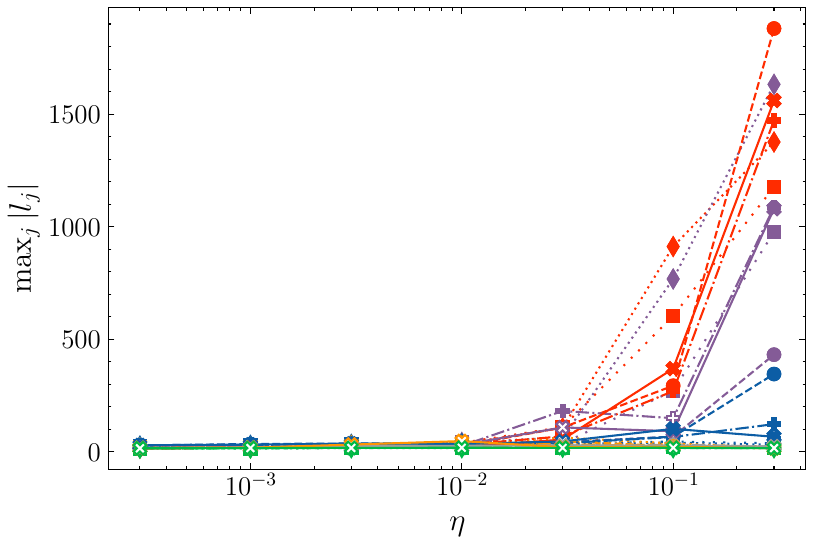} 
    \captionof{figure}{Additional results from the experiments without weight tying. The plots show the dependency of the mean embedding norm (top left), the logits mean (top right), logits standard deviation (bottom left) and maximum absolute logit (bottom right) on the learning rate.}
    \label{fig:wortsman_additional}
\end{minipage} \\ \vspace{3mm}
\begin{minipage}[b]{.98\linewidth}
    \centering
    \hspace{1mm}
    \includegraphics[scale=0.5]{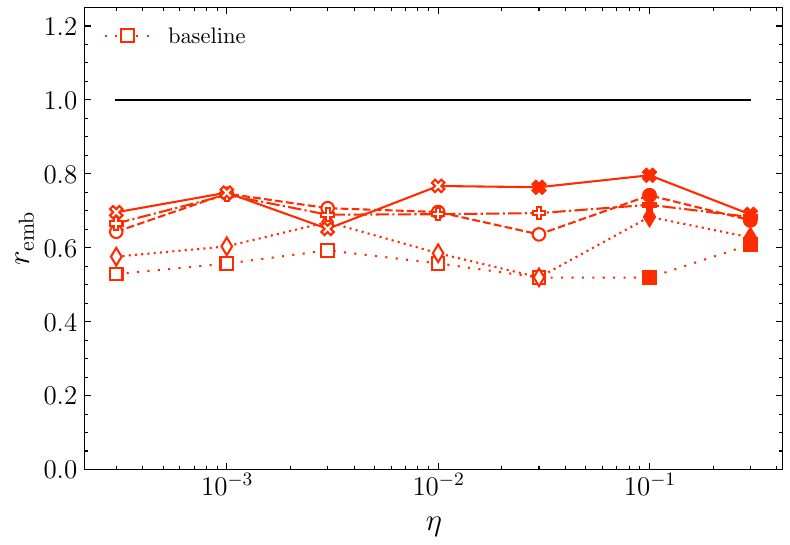} 
    \hspace{3mm}
    \includegraphics[scale=0.5]{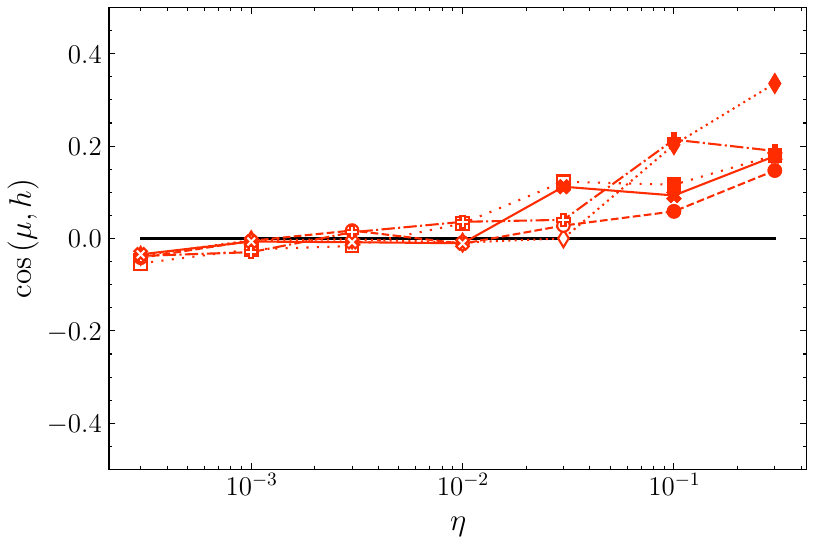} \\ 
    \vspace{2mm} 
    \hspace{1mm}
    \includegraphics[scale=0.5]{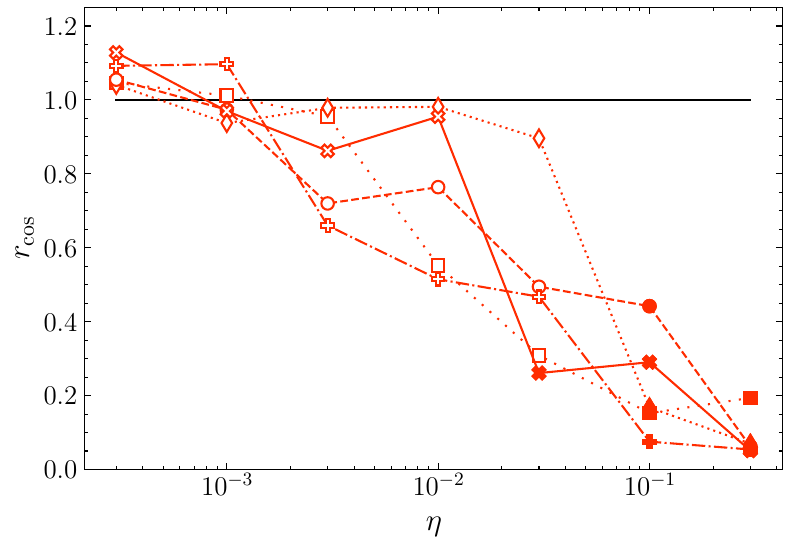} 
    \hspace{5mm}
    \includegraphics[scale=0.5]{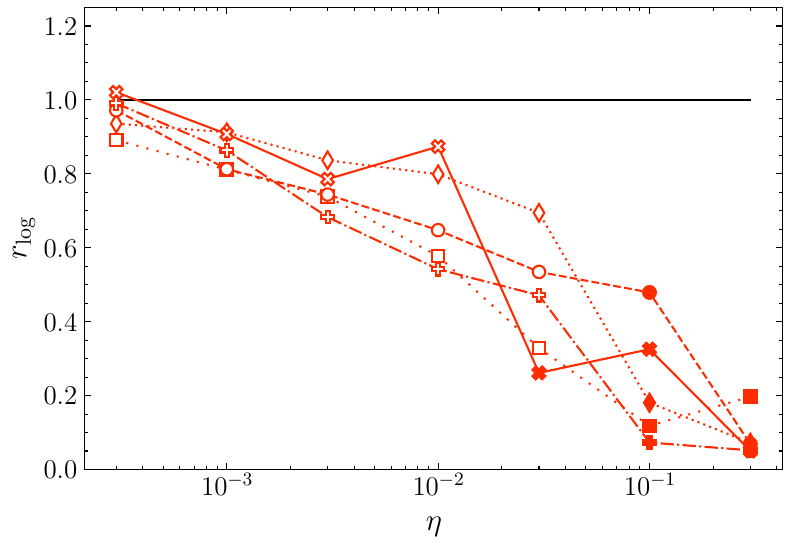} 
    \captionof{figure}{Quantities related to Fact~\ref{lemma_global_logit_bounds}, for all baseline models without weight tying. The different markers and lines represent the different model sizes as in Fig.~\ref{fig:wortsman}, while filled markers indicate divergence. The top left, bottom left and bottom right panels display the averages of $\ratioembeddingi$, $\ratiocosine$ and $\ratiologit$, respectively. 
    The top right panel shows the cosine similarity of the mean embedding $\meanembedding$ and the final hidden state $h$.}
    \label{fig:wortsman_additional_2}
\end{minipage}
\end{figure*}

Our results for the additional quantities mentioned at the end of Sec.~\ref{sec:experiments} are visualized in Fig.~\ref{fig:wortsman_additional} and Fig.~\ref{fig:wortsman_additional_2}. In the following, we will discuss them one by one. In order to keep things simple, we restrict ourselves here to averages over the vocabulary for $\ratioembeddingi$, and averages over the sampled final hidden states or logits for the quantities to which it applies. Their statistical distributions can be found in App.~\ref{app:detailed_results}.  

\paragraph{Mean Embedding}

The mean embedding norm is shown on the top left in Fig.~\ref{fig:wortsman_additional}. As expected, $\mu$-centering maintains a mean embedding of zero, while $\mu$-loss constrains the mean embedding norm to relatively small values. In contrast, both the baseline and the other mitigation strategies lead to large shifts of the mean embedding at at higher learning rates, reflecting the fact that they fail to prevent anisotropic embeddings.

\paragraph{Logit Mean and Standard Deviation (Fact 1 and 2)}

Firstly, regarding the logits mean shown in the top right panel of Fig.~\ref{fig:wortsman_additional}, we find that $\mu$-centering and $\mu$-loss center the logits at and around 0, respectively. This is in accordance with Fact~\ref{lemma_mean_logit}. Similarly, z-loss and logit soft-capping indirectly control the logits mean, although at negative values\footnote{Notably, the logit mean for z-loss is in compliance with theoretical expectations, see App.~\ref{app:zloss} and Eq.~\ref{eq:zloss_minimization_uniform}.}. In contrast, the logits mean diverges at higher learning rates for the baseline, in compliance with the loss divergence observed in Fig.~\ref{fig:wortsman}. The same holds for weight decay, although the divergences are slightly less pronounced.
Secondly, the standard deviation, shown at the bottom left of Fig.~\ref{fig:wortsman}, is the same for $\mu$-centering and the baseline apart from slight statistical differences, at least for lower learning rates for which the baseline training converges. This is consistent with the theoretical prediction, see Fact~\ref{lemma_mec}. In contrast, weight decay, z-loss and $\mu$-loss---since they are regularization methods---change the logit standard deviation slightly. The same holds for logit soft-capping, as it involves a non-linear transformation.

\paragraph{Maximum Absolute Logit (Fact 3)}

Finally, as shown in the bottom right panel of Fig.~\ref{fig:wortsman_additional} and predicted by Fact~\ref{lemma_global_logit_bounds}, both $\mu$-centering and $\mu$-loss restrict the maximum absolute logit to relatively small values around zero. Logit soft-capping achieves the same by design. z-loss also implicitly restricts the maximum logit, albeit to a lesser degree than our methods, which explains the divergence observed for training using z-loss. In contrast, the maximum logit grows very large for the baseline models and, to slightly lesser extent, models using weight decay.
We finish our discussion by studying the ratios $\ratioembeddingi$, $\ratiocosine$, $\ratiologit$ for the baseline models, in order to verify that they fulfill the conditions specified in Fact~\ref{lemma_global_logit_bounds}. This is indeed the case, as the results in Fig.~\ref{fig:wortsman_additional_2} show. In particular, it is obvious that $\ratiocosine$ is the decisive factor leading to $\ratiologit < 1$. We also find confirmation for the alignment of the mean $\meanembedding$ of strongly anisotropic embeddings and the final hidden states $h$ (see Fig.~\ref{fig:mu_centering_alignment}), as expressed by $\cos \left( \meanembedding, \h \right) \neq 0$, for training runs that are prone to divergence.

In summary, we find that our results are in accordance with the theoretical predictions from Sec.~\ref{sec:theory}.

\section{Conclusions}
\label{sec:conclusions}

This paper establishes a link between the problems of anisotropic embeddings and output logit divergence. We have identified the former as as an important factor contributing to the latter, and introduced output embedding centering (OEC) as a theoretically well-founded mitigation strategy. It can be implemented either in terms of the deterministic, hyperparameter-free method $\mu$-centering, or the regularization method $\mu$-loss.
We have conducted systematic experiments comparing these methods to the three most widely used existing mitigation strategies, weight decay, z-loss and logit soft-capping, in a small-scale setting that serves as a proxy for training stability at large. The results show that our methods outperform weight decay and z-loss, while being on par with logit soft-capping. However, in contrast to the latter, our methods address the cause of the problem and avoid model interventions as well as side effects.
The code to reproduce our results is available at \href{https://github.com/flxst/output-embedding-centering}{\nolinkurl{github.com/flxst/output-embedding-centering}}~.

\section*{Acknowledgements}

We would like to thank Amaru Cuba Gyllensten for valuable discussions and feedback.

\bibliography{main}

@article{radford2018improving,
  title={Improving language understanding by generative pre-training},
  author={Radford, Alec and Narasimhan, Karthik and Salimans, Tim and Sutskever, Ilya and others},
  year={2018},
  publisher={San Francisco, CA, USA}
}

@inproceedings{Glorot2010UnderstandingTD,
  title={Understanding the difficulty of training deep feedforward neural networks},
  author={Xavier Glorot and Yoshua Bengio},
  booktitle={International Conference on Artificial Intelligence and Statistics},
  year={2010},
  url={https://api.semanticscholar.org/CorpusID:5575601}
}

@article{penedo2024fineweb,
  title={The fineweb datasets: Decanting the web for the finest text data at scale},
  author={Penedo, Guilherme and Kydl{\'\i}{\v{c}}ek, Hynek and Lozhkov, Anton and Mitchell, Margaret and Raffel, Colin A and Von Werra, Leandro and Wolf, Thomas and others},
  journal={Advances in Neural Information Processing Systems},
  volume={37},
  pages={30811--30849},
  year={2024}
}

@article{vaswani2017attention,
  title={Attention is all you need},
  author={Vaswani, Ashish and Shazeer, Noam and Parmar, Niki and Uszkoreit, Jakob and Jones, Llion and Gomez, Aidan N and Kaiser, {\L}ukasz and Polosukhin, Illia},
  journal={Advances in neural information processing systems},
  volume={30},
  year={2017}
}

@InProceedings{bis2021tmic,
  author =  {Daniel Biś and Maksim Podkorytov and Xiuwen Liu},
  title =   {Too Much in Common: Shifting of Embeddings in Transformer Language Models and its Implications},
  year =    {2021},  
  booktitle = {North American Chapter of the Association for Computational Linguistics (NAACL)},  
}

@article{shazeer2020glu,
  title={Glu variants improve transformer},
  author={Shazeer, Noam},
  journal={arXiv preprint arXiv:2002.05202},
  year={2020}
}

@inproceedings{zhang-etal-2020-revisiting,
    title = "Revisiting Representation Degeneration Problem in Language Modeling",
    author = "Zhang, Zhong  and
      Gao, Chongming  and
      Xu, Cong  and
      Miao, Rui  and
      Yang, Qinli  and
      Shao, Junming",
    editor = "Cohn, Trevor  and
      He, Yulan  and
      Liu, Yang",
    booktitle = "Findings of the Association for Computational Linguistics: EMNLP 2020",
    month = nov,
    year = "2020",
    address = "Online",
    publisher = "Association for Computational Linguistics",
    url = "https://aclanthology.org/2020.findings-emnlp.46",
    doi = "10.18653/v1/2020.findings-emnlp.46",
    pages = "518--527",
}

@misc{gao2019representationdegenerationproblemtraining,
      title={Representation Degeneration Problem in Training Natural Language Generation Models}, 
      author={Jun Gao and Di He and Xu Tan and Tao Qin and Liwei Wang and Tie-Yan Liu},
      year={2019},
      eprint={1907.12009},
      archivePrefix={arXiv},
      primaryClass={cs.CL},
      url={https://arxiv.org/abs/1907.12009}, 
}

@inproceedings{Wang2020ImprovingNL,
  title={Improving Neural Language Generation with Spectrum Control},
  author={Lingxiao Wang and Jing Huang and Kevin Huang and Ziniu Hu and Guangtao Wang and Quanquan Gu},
  booktitle={International Conference on Learning Representations},
  year={2020},
  url={https://api.semanticscholar.org/CorpusID:211145667}
}

@inproceedings{machina-mercer-2024-anisotropy,
    title = "Anisotropy is Not Inherent to Transformers",
    author = "Machina, Anemily  and
      Mercer, Robert",
    editor = "Duh, Kevin  and
      Gomez, Helena  and
      Bethard, Steven",
    booktitle = "Proceedings of the 2024 Conference of the North American Chapter of the Association for Computational Linguistics: Human Language Technologies (Volume 1: Long Papers)",
    month = jun,
    year = "2024",
    address = "Mexico City, Mexico",
    publisher = "Association for Computational Linguistics",
    url = "https://aclanthology.org/2024.naacl-long.274",
    doi = "10.18653/v1/2024.naacl-long.274",
    pages = "4892--4907",
}

@inproceedings{press-wolf-2017-using,
    title = "Using the Output Embedding to Improve Language Models",
    author = "Press, Ofir  and
      Wolf, Lior",
    editor = "Lapata, Mirella  and
      Blunsom, Phil  and
      Koller, Alexander",
    booktitle = "Proceedings of the 15th Conference of the {E}uropean Chapter of the Association for Computational Linguistics: Volume 2, Short Papers",
    month = apr,
    year = "2017",
    address = "Valencia, Spain",
    publisher = "Association for Computational Linguistics",
    url = "https://aclanthology.org/E17-2025/",
    pages = "157--163"
}

@inproceedings{stollenwerk-stollenwerk-2025-better,
    title = "Better Embeddings with Coupled {A}dam",
    author = "Stollenwerk, Felix  and
      Stollenwerk, Tobias",
    editor = "Che, Wanxiang  and
      Nabende, Joyce  and
      Shutova, Ekaterina  and
      Pilehvar, Mohammad Taher",
    booktitle = "Proceedings of the 63rd Annual Meeting of the Association for Computational Linguistics (Volume 1: Long Papers)",
    month = jul,
    year = "2025",
    address = "Vienna, Austria",
    publisher = "Association for Computational Linguistics",
    url = "https://aclanthology.org/2025.acl-long.1321/",
    doi = "10.18653/v1/2025.acl-long.1321",
    pages = "27219--27236",
    ISBN = "979-8-89176-251-0"
}

@misc{chowdhery2022palmscalinglanguagemodeling,
      title={PaLM: Scaling Language Modeling with Pathways}, 
      author={Aakanksha Chowdhery and Sharan Narang and Jacob Devlin and Maarten Bosma and Gaurav Mishra and Adam Roberts and Paul Barham and Hyung Won Chung and Charles Sutton and Sebastian Gehrmann and Parker Schuh and Kensen Shi and Sasha Tsvyashchenko and Joshua Maynez and Abhishek Rao and Parker Barnes and Yi Tay and Noam Shazeer and Vinodkumar Prabhakaran and Emily Reif and Nan Du and Ben Hutchinson and Reiner Pope and James Bradbury and Jacob Austin and Michael Isard and Guy Gur-Ari and Pengcheng Yin and Toju Duke and Anselm Levskaya and Sanjay Ghemawat and Sunipa Dev and Henryk Michalewski and Xavier Garcia and Vedant Misra and Kevin Robinson and Liam Fedus and Denny Zhou and Daphne Ippolito and David Luan and Hyeontaek Lim and Barret Zoph and Alexander Spiridonov and Ryan Sepassi and David Dohan and Shivani Agrawal and Mark Omernick and Andrew M. Dai and Thanumalayan Sankaranarayana Pillai and Marie Pellat and Aitor Lewkowycz and Erica Moreira and Rewon Child and Oleksandr Polozov and Katherine Lee and Zongwei Zhou and Xuezhi Wang and Brennan Saeta and Mark Diaz and Orhan Firat and Michele Catasta and Jason Wei and Kathy Meier-Hellstern and Douglas Eck and Jeff Dean and Slav Petrov and Noah Fiedel},
      year={2022},
      eprint={2204.02311},
      archivePrefix={arXiv},
      primaryClass={cs.CL},
      url={https://arxiv.org/abs/2204.02311}, 
}

@misc{wortsman2023smallscaleproxieslargescaletransformer,
      title={Small-scale proxies for large-scale Transformer training instabilities}, 
      author={Mitchell Wortsman and Peter J. Liu and Lechao Xiao and Katie Everett and Alex Alemi and Ben Adlam and John D. Co-Reyes and Izzeddin Gur and Abhishek Kumar and Roman Novak and Jeffrey Pennington and Jascha Sohl-dickstein and Kelvin Xu and Jaehoon Lee and Justin Gilmer and Simon Kornblith},
      year={2023},
      eprint={2309.14322},
      archivePrefix={arXiv},
      primaryClass={cs.LG},
      url={https://arxiv.org/abs/2309.14322}, 
}

@misc{porian2025resolvingdiscrepanciescomputeoptimalscaling,
 author = {Porian, Tomer and Wortsman, Mitchell and Jitsev, Jenia and Schmidt, Ludwig and Carmon, Yair},
 booktitle = {Advances in Neural Information Processing Systems},
 editor = {A. Globerson and L. Mackey and D. Belgrave and A. Fan and U. Paquet and J. Tomczak and C. Zhang},
 pages = {100535--100570},
 publisher = {Curran Associates, Inc.},
 title = {Resolving Discrepancies in Compute-Optimal Scaling of Language Models},
 url = {https://proceedings.neurips.cc/paper_files/paper/2024/file/b6341525cd84f3be0ef203e4d7cd8556-Paper-Conference.pdf},
 volume = {37},
 year = {2024}
}

@misc{olmo20252olmo2furious,
      title={2 OLMo 2 Furious}, 
      author={{Team OLMo} and Pete Walsh and Luca Soldaini and Dirk Groeneveld and Kyle Lo and Shane Arora and Akshita Bhagia and Yuling Gu and Shengyi Huang and Matt Jordan and Nathan Lambert and Dustin Schwenk and Oyvind Tafjord and Taira Anderson and David Atkinson and Faeze Brahman and Christopher Clark and Pradeep Dasigi and Nouha Dziri and Allyson Ettinger and Michal Guerquin and David Heineman and Hamish Ivison and Pang Wei Koh and Jiacheng Liu and Saumya Malik and William Merrill and Lester James V. Miranda and Jacob Morrison and Tyler Murray and Crystal Nam and Jake Poznanski and Valentina Pyatkin and Aman Rangapur and Michael Schmitz and Sam Skjonsberg and David Wadden and Christopher Wilhelm and Michael Wilson and Luke Zettlemoyer and Ali Farhadi and Noah A. Smith and Hannaneh Hajishirzi},
      year={2025},
      eprint={2501.00656},
      archivePrefix={arXiv},
      primaryClass={cs.CL},
      url={https://arxiv.org/abs/2501.00656}, 
}

@misc{chameleonteam2025chameleonmixedmodalearlyfusionfoundation,
      title={Chameleon: Mixed-Modal Early-Fusion Foundation Models}, 
      author={{Chameleon Team}},
      year={2025},
      eprint={2405.09818},
      archivePrefix={arXiv},
      primaryClass={cs.CL},
      url={https://arxiv.org/abs/2405.09818}, 
}

@inproceedings{Radford2019LanguageMA,
  title={Language Models are Unsupervised Multitask Learners},
  author={Alec Radford and Jeff Wu and Rewon Child and David Luan and Dario Amodei and Ilya Sutskever},
  year={2019},
  url={https://api.semanticscholar.org/CorpusID:160025533}
}

@InProceedings{pmlr-v162-wang22u,
  title = 	 {What Language Model Architecture and Pretraining Objective Works Best for Zero-Shot Generalization?},
  author =       {Wang, Thomas and Roberts, Adam and Hesslow, Daniel and Scao, Teven Le and Chung, Hyung Won and Beltagy, Iz and Launay, Julien and Raffel, Colin},
  booktitle = 	 {Proceedings of the 39th International Conference on Machine Learning},
  pages = 	 {22964--22984},
  year = 	 {2022},
  editor = 	 {Chaudhuri, Kamalika and Jegelka, Stefanie and Song, Le and Szepesvari, Csaba and Niu, Gang and Sabato, Sivan},
  volume = 	 {162},
  series = 	 {Proceedings of Machine Learning Research},
  month = 	 {17--23 Jul},
  publisher =    {PMLR},
  url = 	 {https://proceedings.mlr.press/v162/wang22u.html}
}

@techreport{OLMo3_2025,
  title       = {OLMo 3: Charting a path through the model flow to lead open-source AI},
  author      = {{Team OLMo} and {Allen Institute for AI}},
  year        = {2025},
  month       = nov,
  institution = {Allen Institute for AI},
  url         = {https://allenai.org/blog/olmo3},
  type        = {Technical Report}
}

@INPROCEEDINGS{10189242,
  author={Jiang, Zixuan and Gu, Jiaqi and Pan, David Z.},
  booktitle={2023 IEEE International Conference on Omni-layer Intelligent Systems (COINS)}, 
  title={NormSoftmax: Normalizing the Input of Softmax to Accelerate and Stabilize Training}, 
  year={2023},
  volume={},
  number={},
  pages={1-6},
  doi={10.1109/COINS57856.2023.10189242}
}

@misc{gemmateam2024gemma2improvingopen,
      title={Gemma 2: Improving Open Language Models at a Practical Size}, 
      author={{Gemma Team} and Morgane Riviere and Shreya Pathak and Pier Giuseppe Sessa and Cassidy Hardin and Surya Bhupatiraju and Léonard Hussenot and Thomas Mesnard and Bobak Shahriari and Alexandre Ramé and Johan Ferret and Peter Liu and Pouya Tafti and Abe Friesen and Michelle Casbon and Sabela Ramos and Ravin Kumar and Charline Le Lan and Sammy Jerome and Anton Tsitsulin and Nino Vieillard and Piotr Stanczyk and Sertan Girgin and Nikola Momchev and Matt Hoffman and Shantanu Thakoor and Jean-Bastien Grill and Behnam Neyshabur and Olivier Bachem and Alanna Walton and Aliaksei Severyn and Alicia Parrish and Aliya Ahmad and Allen Hutchison and Alvin Abdagic and Amanda Carl and Amy Shen and Andy Brock and Andy Coenen and Anthony Laforge and Antonia Paterson and Ben Bastian and Bilal Piot and Bo Wu and Brandon Royal and Charlie Chen and Chintu Kumar and Chris Perry and Chris Welty and Christopher A. Choquette-Choo and Danila Sinopalnikov and David Weinberger and Dimple Vijaykumar and Dominika Rogozińska and Dustin Herbison and Elisa Bandy and Emma Wang and Eric Noland and Erica Moreira and Evan Senter and Evgenii Eltyshev and Francesco Visin and Gabriel Rasskin and Gary Wei and Glenn Cameron and Gus Martins and Hadi Hashemi and Hanna Klimczak-Plucińska and Harleen Batra and Harsh Dhand and Ivan Nardini and Jacinda Mein and Jack Zhou and James Svensson and Jeff Stanway and Jetha Chan and Jin Peng Zhou and Joana Carrasqueira and Joana Iljazi and Jocelyn Becker and Joe Fernandez and Joost van Amersfoort and Josh Gordon and Josh Lipschultz and Josh Newlan and Ju-yeong Ji and Kareem Mohamed and Kartikeya Badola and Kat Black and Katie Millican and Keelin McDonell and Kelvin Nguyen and Kiranbir Sodhia and Kish Greene and Lars Lowe Sjoesund and Lauren Usui and Laurent Sifre and Lena Heuermann and Leticia Lago and Lilly McNealus and Livio Baldini Soares and Logan Kilpatrick and Lucas Dixon and Luciano Martins and Machel Reid and Manvinder Singh and Mark Iverson and Martin Görner and Mat Velloso and Mateo Wirth and Matt Davidow and Matt Miller and Matthew Rahtz and Matthew Watson and Meg Risdal and Mehran Kazemi and Michael Moynihan and Ming Zhang and Minsuk Kahng and Minwoo Park and Mofi Rahman and Mohit Khatwani and Natalie Dao and Nenshad Bardoliwalla and Nesh Devanathan and Neta Dumai and Nilay Chauhan and Oscar Wahltinez and Pankil Botarda and Parker Barnes and Paul Barham and Paul Michel and Pengchong Jin and Petko Georgiev and Phil Culliton and Pradeep Kuppala and Ramona Comanescu and Ramona Merhej and Reena Jana and Reza Ardeshir Rokni and Rishabh Agarwal and Ryan Mullins and Samaneh Saadat and Sara Mc Carthy and Sarah Cogan and Sarah Perrin and Sébastien M. R. Arnold and Sebastian Krause and Shengyang Dai and Shruti Garg and Shruti Sheth and Sue Ronstrom and Susan Chan and Timothy Jordan and Ting Yu and Tom Eccles and Tom Hennigan and Tomas Kocisky and Tulsee Doshi and Vihan Jain and Vikas Yadav and Vilobh Meshram and Vishal Dharmadhikari and Warren Barkley and Wei Wei and Wenming Ye and Woohyun Han and Woosuk Kwon and Xiang Xu and Zhe Shen and Zhitao Gong and Zichuan Wei and Victor Cotruta and Phoebe Kirk and Anand Rao and Minh Giang and Ludovic Peran and Tris Warkentin and Eli Collins and Joelle Barral and Zoubin Ghahramani and Raia Hadsell and D. Sculley and Jeanine Banks and Anca Dragan and Slav Petrov and Oriol Vinyals and Jeff Dean and Demis Hassabis and Koray Kavukcuoglu and Clement Farabet and Elena Buchatskaya and Sebastian Borgeaud and Noah Fiedel and Armand Joulin and Kathleen Kenealy and Robert Dadashi and Alek Andreev},
      year={2024},
      eprint={2408.00118},
      archivePrefix={arXiv},
      primaryClass={cs.CL},
      url={https://arxiv.org/abs/2408.00118}, 
}

@misc{yang2025baichuan2openlargescale,
      title={Baichuan 2: Open Large-scale Language Models}, 
      author={Aiyuan Yang and Bin Xiao and Bingning Wang and Borong Zhang and Ce Bian and Chao Yin and Chenxu Lv and Da Pan and Dian Wang and Dong Yan and Fan Yang and Fei Deng and Feng Wang and Feng Liu and Guangwei Ai and Guosheng Dong and Haizhou Zhao and Hang Xu and Haoze Sun and Hongda Zhang and Hui Liu and Jiaming Ji and Jian Xie and JunTao Dai and Kun Fang and Lei Su and Liang Song and Lifeng Liu and Liyun Ru and Luyao Ma and Mang Wang and Mickel Liu and MingAn Lin and Nuolan Nie and Peidong Guo and Ruiyang Sun and Tao Zhang and Tianpeng Li and Tianyu Li and Wei Cheng and Weipeng Chen and Xiangrong Zeng and Xiaochuan Wang and Xiaoxi Chen and Xin Men and Xin Yu and Xuehai Pan and Yanjun Shen and Yiding Wang and Yiyu Li and Youxin Jiang and Yuchen Gao and Yupeng Zhang and Zenan Zhou and Zhiying Wu},
      year={2025},
      eprint={2309.10305},
      archivePrefix={arXiv},
      primaryClass={cs.CL},
      url={https://arxiv.org/abs/2309.10305}, 
}

@inproceedings{
takase2025spike,
title={Spike No More: Stabilizing the Pre-training of Large Language Models},
author={Sho Takase and Shun Kiyono and Sosuke Kobayashi and Jun Suzuki},
booktitle={Second Conference on Language Modeling},
year={2025},
url={https://openreview.net/forum?id=52YBEzcI0l}
}

@InProceedings{pmlr-v202-dehghani23a,
  title = 	 {Scaling Vision Transformers to 22 Billion Parameters},
  author =       {Dehghani, Mostafa and Djolonga, Josip and Mustafa, Basil and Padlewski, Piotr and Heek, Jonathan and Gilmer, Justin and Steiner, Andreas Peter and Caron, Mathilde and Geirhos, Robert and Alabdulmohsin, Ibrahim and Jenatton, Rodolphe and Beyer, Lucas and Tschannen, Michael and Arnab, Anurag and Wang, Xiao and Riquelme Ruiz, Carlos and Minderer, Matthias and Puigcerver, Joan and Evci, Utku and Kumar, Manoj and Steenkiste, Sjoerd Van and Elsayed, Gamaleldin Fathy and Mahendran, Aravindh and Yu, Fisher and Oliver, Avital and Huot, Fantine and Bastings, Jasmijn and Collier, Mark and Gritsenko, Alexey A. and Birodkar, Vighnesh and Vasconcelos, Cristina Nader and Tay, Yi and Mensink, Thomas and Kolesnikov, Alexander and Pavetic, Filip and Tran, Dustin and Kipf, Thomas and Lucic, Mario and Zhai, Xiaohua and Keysers, Daniel and Harmsen, Jeremiah J. and Houlsby, Neil},
  booktitle = 	 {Proceedings of the 40th International Conference on Machine Learning},
  pages = 	 {7480--7512},
  year = 	 {2023},
  editor = 	 {Krause, Andreas and Brunskill, Emma and Cho, Kyunghyun and Engelhardt, Barbara and Sabato, Sivan and Scarlett, Jonathan},
  volume = 	 {202},
  series = 	 {Proceedings of Machine Learning Research},
  month = 	 {23--29 Jul},
  publisher =    {PMLR},
  url = 	 {https://proceedings.mlr.press/v202/dehghani23a.html}
}

@article{arora-etal-2016-latent,
    title = "A Latent Variable Model Approach to {PMI}-based Word Embeddings",
    author = "Arora, Sanjeev  and
      Li, Yuanzhi  and
      Liang, Yingyu  and
      Ma, Tengyu  and
      Risteski, Andrej",
    editor = "Lee, Lillian  and
      Johnson, Mark  and
      Toutanova, Kristina",
    journal = "Transactions of the Association for Computational Linguistics",
    volume = "4",
    year = "2016",
    address = "Cambridge, MA",
    publisher = "MIT Press",
    url = "https://aclanthology.org/Q16-1028/",
    doi = "10.1162/tacl_a_00106",
    pages = "385--399"
}

@misc{mu2018allbutthetopsimpleeffectivepostprocessing,
      title={All-but-the-Top: Simple and Effective Postprocessing for Word Representations}, 
      author={Jiaqi Mu and Suma Bhat and Pramod Viswanath},
      year={2018},
      eprint={1702.01417},
      archivePrefix={arXiv},
      primaryClass={cs.CL},
      url={https://arxiv.org/abs/1702.01417}, 
}

@misc{zoph2022stmoedesigningstabletransferable,
      title={ST-MoE: Designing Stable and Transferable Sparse Expert Models}, 
      author={Barret Zoph and Irwan Bello and Sameer Kumar and Nan Du and Yanping Huang and Jeff Dean and Noam Shazeer and William Fedus},
      year={2022},
      eprint={2202.08906},
      archivePrefix={arXiv},
      primaryClass={cs.CL},
      url={https://arxiv.org/abs/2202.08906}, 
}

@inproceedings{
loshchilov2018decoupled,
title={Decoupled Weight Decay Regularization},
author={Ilya Loshchilov and Frank Hutter},
booktitle={International Conference on Learning Representations},
year={2019},
url={https://openreview.net/forum?id=Bkg6RiCqY7},
}
\bibliographystyle{tmlr}

\clearpage
\appendix
\section{Hyperparameters}
\label{app:hyperparameters}

All our experiments use the architecture and hyperparameters specified in Tab.~\ref{tab:model_architecture}.
\begin{table}[!ht]
\centering
\scriptsize
\caption{Architectural details and hyperparameters used in all our experiments. All settings match the ones from \citet{wortsman2023smallscaleproxieslargescaletransformer}, with six exceptions. These are highlighted in bold, with the choice from \citet{wortsman2023smallscaleproxieslargescaletransformer} being specified in parentheses.}
\begin{tabular}{ll}
\toprule
optimizer & AdamW \\ 
$\beta_1$ & 0.9 \\
$\beta_2$ & 0.95 \\
$\epsilon$ & 1e-8 \\
weight decay & 0.0 \\
gradient clipping & 1.0 \\
dropout & 0.0 \\
\textbf{weight tying} & \textbf{false \& true (false)} \\
qk-layernorm &  yes \\
bias & no \\
learning rate schedule & cosine decay \\
learning rate minimum & 1e-5 \\
layer normalization & LayerNorm \\
precision & BF16 \\
positional embedding & RoPE \\ 
\textbf{vocab size} & \textbf{50304 (32101)}  \\
\textbf{hidden activation} & \textbf{SwiGLU (GeLU)} \\
\textbf{sequence length} & \textbf{2048 (512)} \\
\textbf{batch size (samples)} & \textbf{64 (256)} \\
batch size (tokens) & 131072 \\
training length & 100000 steps $\approx$ 13.1B tokens \\
warmup & 5000 steps $\approx$ 0.7B tokens \\
embedding initialization & Normal with standard deviation $1/\sqrt{d}$ \\
\textbf{weight initialization} & \textbf{Xavier with average of fan\_in and fan\_out} \\
 & \textbf{(Xavier with fan\_in, truncated)} \\
\bottomrule 
\end{tabular}
\label{tab:model_architecture}
\end{table}

\section{Details on z-loss}
\label{app:zloss}

\begin{figure}[!th]
\centering
\includegraphics[scale=0.6]{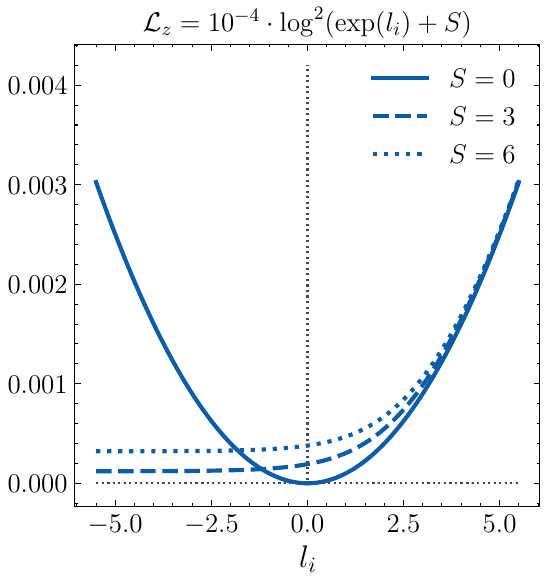}
\caption{Illustration of $\mathcal{L}_z$ as a function of a single logit $l_i$ as given by Eq.~\ref{eq:zloss_function_1D}, for 3 different values of $S = \sum_{j \neq i} \exp{(l_j)}$.}
\label{fig:zloss_illustrations_1D}
\end{figure}

\begin{figure*}[!th]
\centering
\includegraphics[scale=0.8]{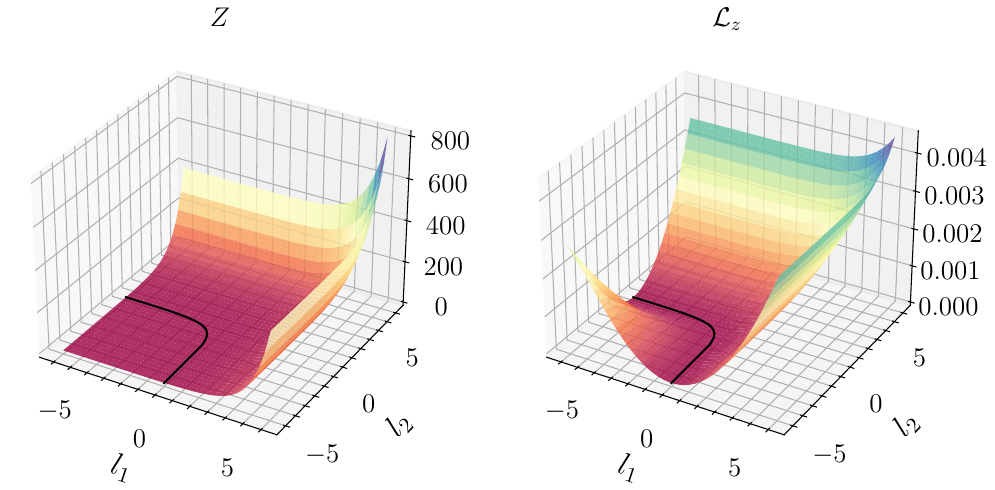}
\caption{The quantities $Z$ (left, Eq.~\ref{eq:Z}) and $\mathcal{L}_z$ (right, Eq.~\ref{eq:zloss}) as a function of two logits, i.e. $V=2$. The black curves represent the z-loss optimum, $Z = 1$ and $\mathcal{L}_z = 0$.}
\label{fig:zloss_illustrations_2D}
\end{figure*}

Fig.~\ref{fig:zloss_illustrations_1D} shows $\mathcal{L}_z$ as a function of a single logit $l_i$, see Eq.~\ref{eq:zloss_function_1D}, for 3 different values of $S = \sum_{j \neq i} \exp{(l_j)}$. 
It illustrates that the function is even only for the theoretical edge case of $S = 0$, corresponding to $\lim l_j \to -\infty$ for all $j \neq i$.
The fact that z-loss treats positive and negative logits differently is also evident from Fig.~\ref{fig:zloss_illustrations_2D}. It shows $Z$ and $\mathcal{L}_z$ for a toy example of two logits, $l_1$ and $l_2$. Neither function is invariant under the transformation $l_i \to -l_i$ (i.e. a $180^\circ$ rotation around the z-axis). In fact, the limit $\mathcal{L}_z \to \infty$, which corresponds to either $Z \to 0$ or $Z \to \infty$, is approached differently: while $Z \to 0$ requires \textit{all} logits to diverge negatively, $Z \to \infty$ is obtained if \textit{any} logit diverges positively.
At the same time, the limit $\mathcal{L}_z \to 0$, which corresponds to $Z \to 1$, requires all logits to be non-positive, $\max_j l_j = 0$, while allowing for the occurrence of very large negative logits, $l_i \to - \infty$. 
For instance, possible logit configurations minimizing z-loss are 
\begin{equation}
    l_i = 0 \qquad \text{and} \qquad l_j = -\infty \quad \forall \ j \neq i
\end{equation}
or
\begin{equation}
    l_j = - \log(V) \quad \forall \ j \in \mathcal{V}
    \label{eq:zloss_minimization_uniform}
\end{equation}
In particular, unlike $\mu$-centering and $\mu$-loss (see Fact~\ref{lemma_mec}), the minimization of z-loss does not enforce $\meanlogit \to 0$.

\section{Hyperparameter Sensitivity}
\label{app:ablations}

\begin{figure*}[!p]
    \centering 
    \captionof{table}{Learning rate sensitivity for $\mu$-loss (left) and z-loss (right) with different regularization hyperparameters $\lambda$ (specified in the column headers).}
    \begin{minipage}{\textwidth}
        \centering
        \hspace{0.0cm}
        \begin{subtable}[t]{0.4\linewidth}
            \centering
            \hspace{0.3cm} $\boldsymbol{\mu}$\textbf{-loss} \\ \vspace{0.7em}
            \scriptsize
            \begin{tabular}{rcccc}
                \multicolumn{5}{c}{LRS ($\downarrow$) ~|~ weight tying = false} \\
                \toprule
                $N$ & $10^{-7}$ & $10^{-4}$ & $10^{-1}$ & $10^{2}$ \\ \midrule
                16M & 0.182 & \textbf{0.031} & \textbf{0.031} & 0.054 \\ 
                29M & 0.052 & \textbf{0.027} & 0.034 & 0.040 \\ 
                57M & 0.110 & \textbf{0.031} & 0.038 & 0.033 \\ 
                109M & 0.125 & 0.046 & 0.048 & \textbf{0.034} \\ 
                221M & 0.129 & 0.056 & 0.056 & \textbf{0.055} \\ \bottomrule
            \end{tabular}
            \vspace{3mm}
            \label{tab:ablations_mer}
        \end{subtable}
        \hspace{1.2cm}
        \begin{subtable}[t]{0.4\linewidth}
            \centering
            \hspace{0em} \textbf{z-loss} \\ \vspace{0.9em}
            \scriptsize
            \begin{tabular}{rcccc}
                \multicolumn{5}{c}{LRS ($\downarrow$) ~|~ weight tying = false} \\
                \toprule
                $N$ & $10^{-7}$ & $10^{-4}$ & $10^{-1}$ & $10^{2}$ \\ \midrule
                16M & 0.037 & 0.054 & \textbf{0.032} & 1.156 \\ 
                29M & 0.044 & \textbf{0.033} & 0.043 & 1.780 \\ 
                57M & 0.107 & 0.235 & \textbf{0.047} & 1.392 \\ 
                109M & 0.076 & 0.118 & \textbf{0.059} & 2.150 \\ 
                221M & 0.131 & 0.109 & \textbf{0.101} & 2.166 \\ \bottomrule
            \end{tabular}
            \label{tab:ablations_z}
        \end{subtable}
        \label{tab:overview_comparison}
    \end{minipage} 
    \\ \vspace{5mm}
    \begin{minipage}{\textwidth}
    \centering
    \hspace{-2em}
    \begin{subtable}[t]{0.4\textwidth}
        \centering
        \hspace{2.5em} $\boldsymbol{\mu}$\textbf{-loss} \\  \vspace{0.3em}
        \includegraphics[scale=0.5]{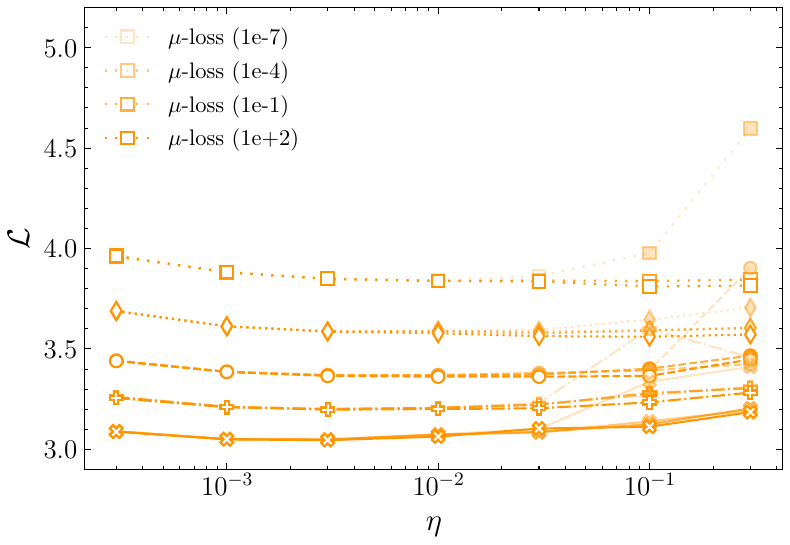}
        \label{fig:wortsman_ablations_E_top}
    \end{subtable}
    \hspace{1.2cm}
    \begin{subtable}[t]{0.4\textwidth}
        \centering
        \hspace{1.7em} \textbf{z-loss} \\ \vspace{0.5em}
        \includegraphics[scale=0.5]{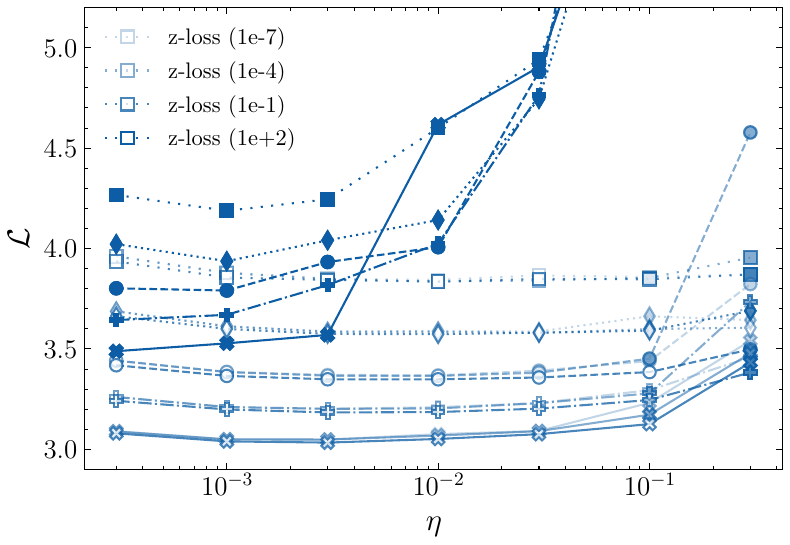}
        \label{fig:wortsman_ablations_z_top}
    \end{subtable} 
    \\
    \hspace{-2.5em}
    \vspace{-0.5em} 
    \begin{subtable}[t]{0.4\textwidth}
        \centering
        \includegraphics[scale=0.5]{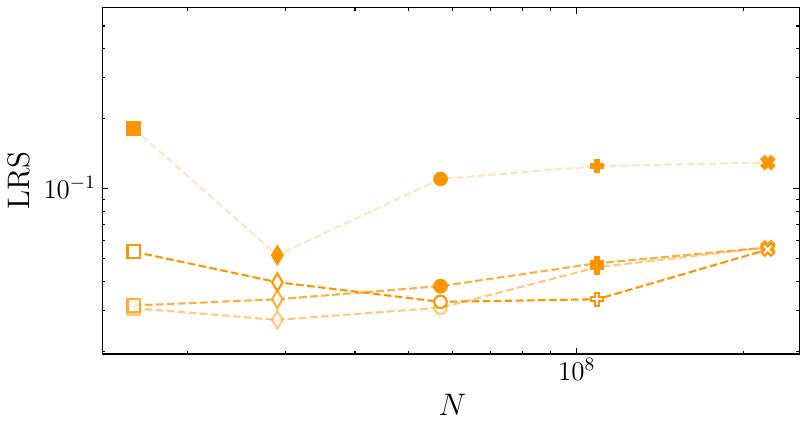}
        \label{fig:wortsman_ablations_E_bottom}
    \end{subtable}
    \hspace{1.2cm}
    \begin{subtable}[t]{0.4\textwidth}
        \centering
        \includegraphics[scale=0.5]{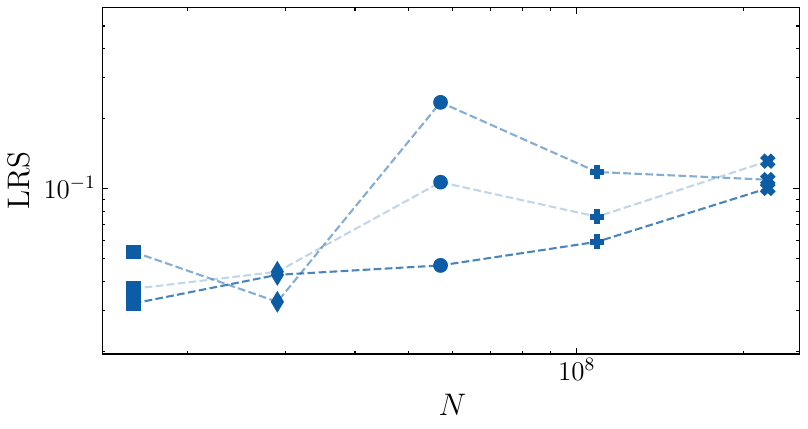}
        \label{fig:wortsman_ablations_z_bottom}
    \end{subtable}
    \vspace{-1em}
    \captionof{figure}{Hyperparameter dependency of $\mu$-loss (left) and z-loss (right). The top plots show loss $\mathcal{L}$ vs. learning rate $\eta$, while the bottom plots show learning rate sensitivity vs. model size $N$, as listed in Tab.~\ref{tab:overview_comparison}.}
    \label{fig:ablations}
    \end{minipage}
\end{figure*}

In our experiments in Sec.~\ref{sec:experiments}, the regularization hyperparameters have been set to their default value $\lambda = 10^{-4}$ for both regularization methods, z-loss (cf.~Eq.~\ref{eq:zloss}) and $\mu$-loss (cf.~Eq.~\ref{eq:lambda_default}).
Here, we vary the regularization hyperparameter for those methods,
\begin{align}
\lambda &\in \{ 10^{-7}, 10^{-4}, 10^{-1}, 10^{2} \}
\label{eq:lambda_ablations}
\end{align}
and determine the learning rate sensitivity as in Sec.~\ref{sec:results}, for each choice of $\lambda$.
The results are presented in Tab.~\ref{tab:overview_comparison} and Fig.~\ref{fig:ablations}. 
For $\mu$-loss, hyperparameter tuning is notably straightforward: the regularization coefficient only needs to be sufficiently large to enforce the centering effect. In fact, for larger values ($\lambda \ge 10^{-4}$), the training is stable and does not exhibit a strong dependency on the exact value of $\lambda$. Only when $\lambda$ is too small ($\lambda = 10^{-7}$), we observe that the loss diverges for large learning rates across all model sizes.
This behavior stands in contrast to z-loss, which requires more careful tuning. Severe divergences appear for $\lambda=10^2$, but also for lower values of $\lambda$ in conjunction with large learning rates. 
Our results indicate that the optimal value for z-loss is $\lambda=10^{-1}$, which is significantly larger than the previously assumed optimal value of $10^{-4}$. Importantly, however, even for the optimal $\lambda$, z-loss is outperformed by both $\mu$-loss and $\mu$-centering. This performance gap is evident from the learning rate sensitivity values for the largest model size $N=221$ in Tab.~\ref{tab:overview_comparison}, as well as in the comparison of the rightmost points---corresponding to the largest model size---across the learning rate sensitivity plots in Fig.~\ref{fig:ablations}.

\newpage
\section{Results for Weight Tying}
\label{app:wt}

\begin{figure*}[ht]
\centering
\includegraphics[scale=0.56]{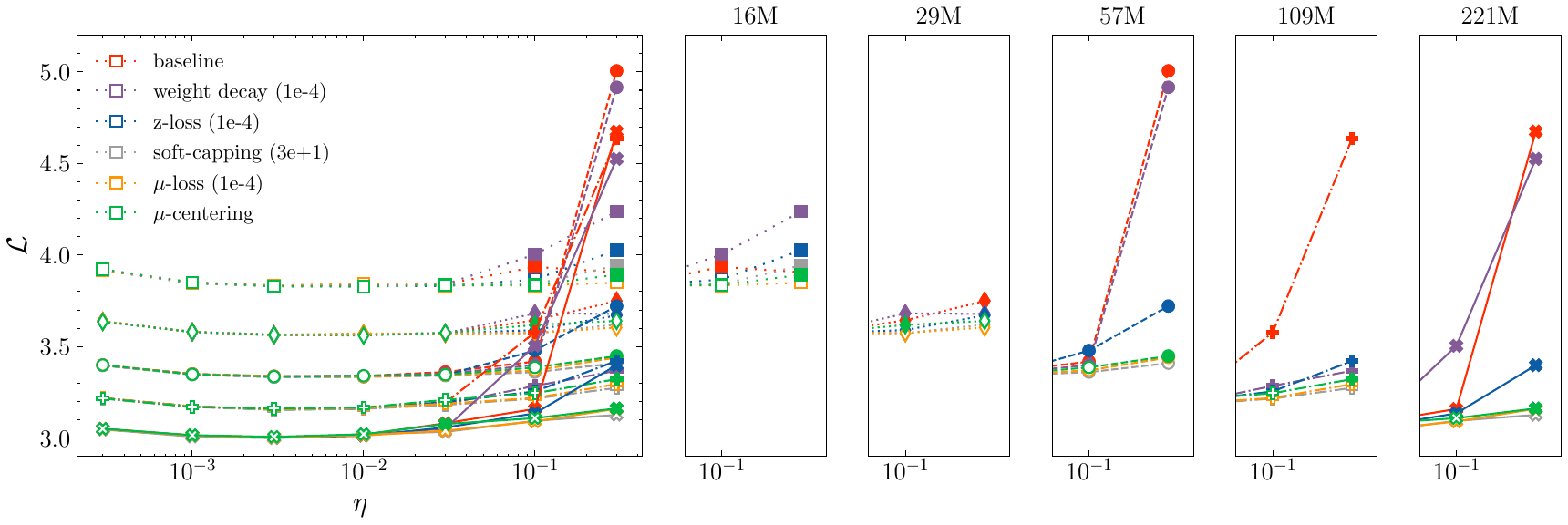}
\captionof{figure}{Dependency of the loss $\mathcal{L}$ on the learning rate $\eta$, for models with weight tying. The panel on the very left shows all results together, while the others focus on the largest two learning rates and a single model size $N$ (specified in the respective title). Filled and empty markers indicate divergence and convergence, respectively.}
\label{fig:wortsman_wt}
\vspace{4mm}
\begin{minipage}[b]{.42\linewidth}
\centering
\includegraphics[scale=0.5]{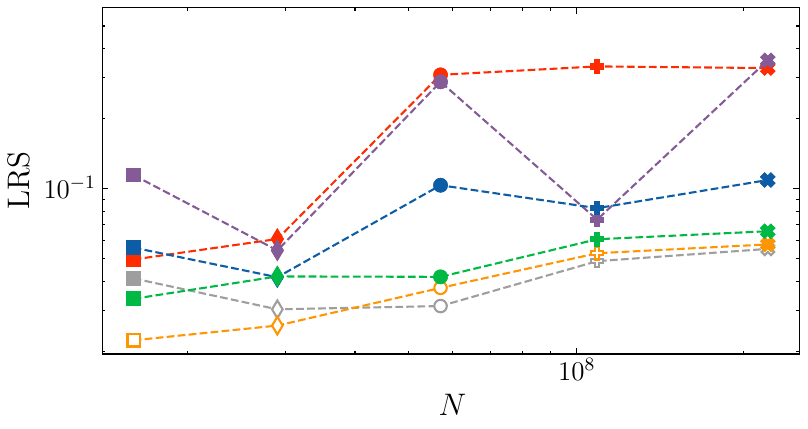}
\captionof{figure}{Learning rate sensitivity LRS for models with weight tying. Filled markers indicate that the training for at least one learning rate $\eta$ diverged. The colors represent the different variants as in Fig.~\ref{fig:wortsman_wt}.}
\label{fig:wortsman_lrs_wt}
\end{minipage}\hfill
\begin{minipage}[b]{.55\linewidth}
\centering
\scriptsize
\captionof{table}{Learning rate sensitivity LRS for models without weight tying. Underlined numbers indicate that the training for all learning rates converged, corresponding to the empty markers in Fig.~\ref{fig:wortsman_lrs_wt}. In addition, the best result for each model size $N$ is highlighted in bold.}
\begin{tabular}{rcccccc}
    \multicolumn{7}{c}{LRS ($\downarrow$) ~|~ weight tying = true} \\
    \toprule
    $N$ & \colorbox{clrbaseline!30}{basel.} & \colorbox{clrweightdecay!30}{w.-d.} & \colorbox{clrzloss!30}{z-loss} & \colorbox{clrsoftcapping!30}{soft-c.} & \colorbox{clrmuloss!30}{$\mu$-loss} & \colorbox{clrmucentering!30}{$\mu$-cent.}  \\ \midrule
    16M & 0.050 & 0.114 & 0.056 & 0.041 & \underline{\textbf{0.022}} & 0.034 \\ 
    29M & 0.061 & 0.054 & 0.042 & \underline{0.030} & \underline{\textbf{0.026}} & 0.042 \\ 
    57M & 0.308 & 0.288 & 0.103 & \underline{\textbf{0.031}} & \underline{0.038} & 0.042 \\ 
    109M & 0.335 & 0.074 & 0.083 & \underline{\textbf{0.049}} & \underline{0.053} & 0.061 \\ 
    221M & 0.330 & 0.356 & 0.109 & \underline{\textbf{0.055}} & 0.058 & 0.066 \\
    \bottomrule
\end{tabular}
\vspace{4mm}
\label{tab:wortsman_lrs_wt}
\end{minipage}
\end{figure*}

The results of our experiments with weight tying are shown in Fig.~\ref{fig:wortsman_wt}, Fig.~\ref{fig:wortsman_lrs_wt} and Tab.~\ref{tab:wortsman_lrs_wt}. They are very similar to their counterparts without weight tying, see Fig.~\ref{fig:wortsman}, Fig.~\ref{fig:wortsman_lrs} and Tab.~\ref{tab:wortsman_lrs}. In particular, the performance ranking is the same as in Eq.~\ref{eq:main_conclusion}. Note that, with regard to our methods, subtracting the mean from the input embeddings $e_i$ amounts to subtracting an effective bias term $\meanembedding$.

\newpage
\section{Computational Overhead}
\label{app:additional_results} 

Tab.~\ref{tab:overview_training_time} lists the additional time $\Delta T$ of a method compared to the baseline to train a model on 4 A100 GPUs using data parallelism. 
\begin{table}[th]
\centering
\scriptsize
\caption{Additional training time $\Delta T$ relative to the baseline, without (left) and with (right) weight tying.}
\begin{tabular}{rccccc}
\multicolumn{6}{c}{$\Delta T$ ($\downarrow$) ~|~ weight tying = false} \\
\toprule
$N$ & \colorbox{clrweightdecay!30}{w.-d.} & \colorbox{clrzloss!30}{z-loss} & \colorbox{clrsoftcapping!30}{soft-c.} & \colorbox{clrmuloss!30}{$\mu$-loss} & \colorbox{clrmucentering!30}{$\mu$-cent.} \\ \midrule
16M & -1.0\% & 6.4\% & 10.0\% & 0.4\% & 0.6\% \\ 
29M & -0.5\% & 4.3\% & 6.4\% & 0.7\% & 0.5\% \\ 
57M & -0.3\% & 2.5\% & 3.9\% & 0.6\% & 0.4\% \\ 
109M & -0.5\% & 1.5\% & 2.0\% & 0.4\% & 0.4\% \\ 
221M & -0.7\% & 0.8\% & 0.7\% & 0.2\% & 0.3\% \\ 
\bottomrule
\end{tabular}
\hspace{3mm}
\begin{tabular}{rccccc}
\multicolumn{6}{c}{$\Delta T$ ($\downarrow$) ~|~ weight tying = true} \\
\toprule
$N$ & \colorbox{clrweightdecay!30}{w.-d.} & \colorbox{clrzloss!30}{z-loss} & \colorbox{clrsoftcapping!30}{soft-c.} & \colorbox{clrmuloss!30}{$\mu$-loss} & \colorbox{clrmucentering!30}{$\mu$-cent.} \\ \midrule
16M & 0.4\% & 6.6\% & 11.2\% & 0.5\% & 0.7\% \\ 
29M & 0.0\% & 4.0\% & 7.1\% & 0.4\% & 0.5\% \\ 
57M & 0.2\% & 2.7\% & 4.5\% & 0.8\% & 0.7\% \\ 
109M & 0.1\% & 1.8\% & 2.8\% & 0.4\% & 0.6\% \\ 
221M & 0.1\% & 1.0\% & 1.4\% & 0.0\% & 0.2\% \\ 
\bottomrule
\end{tabular}
\label{tab:overview_training_time}
\end{table}
It shows that $\mu$-loss and $\mu$-centering only have a very small effect on training time. In particular, the computational overhead is smaller than for logit soft-capping and z-loss. Note that for larger models and training setups that require parallelization, tensor parallelism can be applied to the embedding matrix such that it is split across the hidden dimension (rather than the vocabulary dimension). In that case, the operations for both $\mu$-centering and $\mu$-loss become embarrassingly parallel and require no particular collective communication. In addition, $\mu$-centering merely involves a subtraction, which is computationally negligible. Hence, we expect our methods to be at least as computationally cheap as the other mitigation strategies, also in a large-scale training setting that involves parallelization.

\section{Statistical Distributions of Results}
\label{app:detailed_results}

\begin{figure*}[ht]
    \centering
    \includegraphics[scale=0.48]{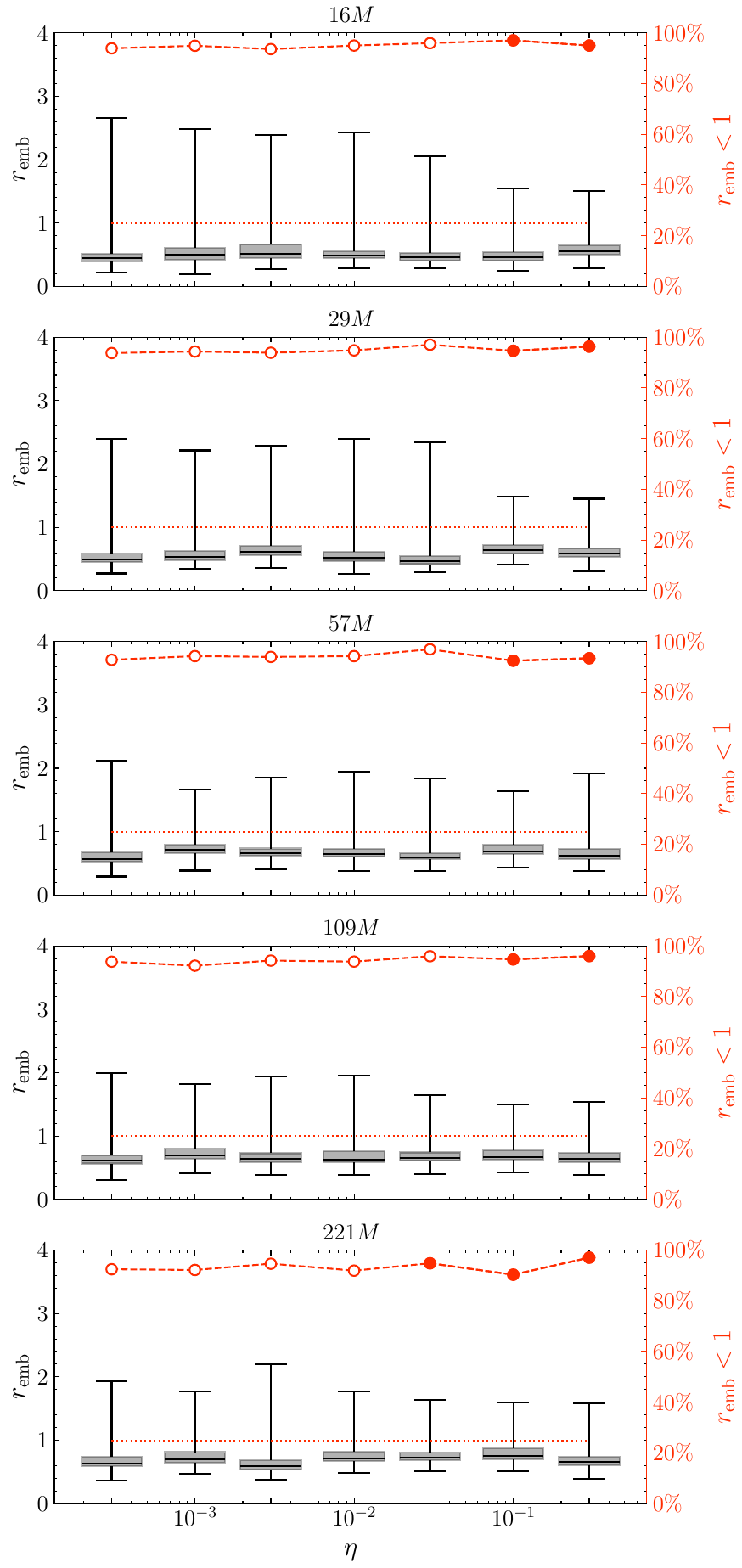}
    \hspace{1cm}
    \includegraphics[scale=0.48]{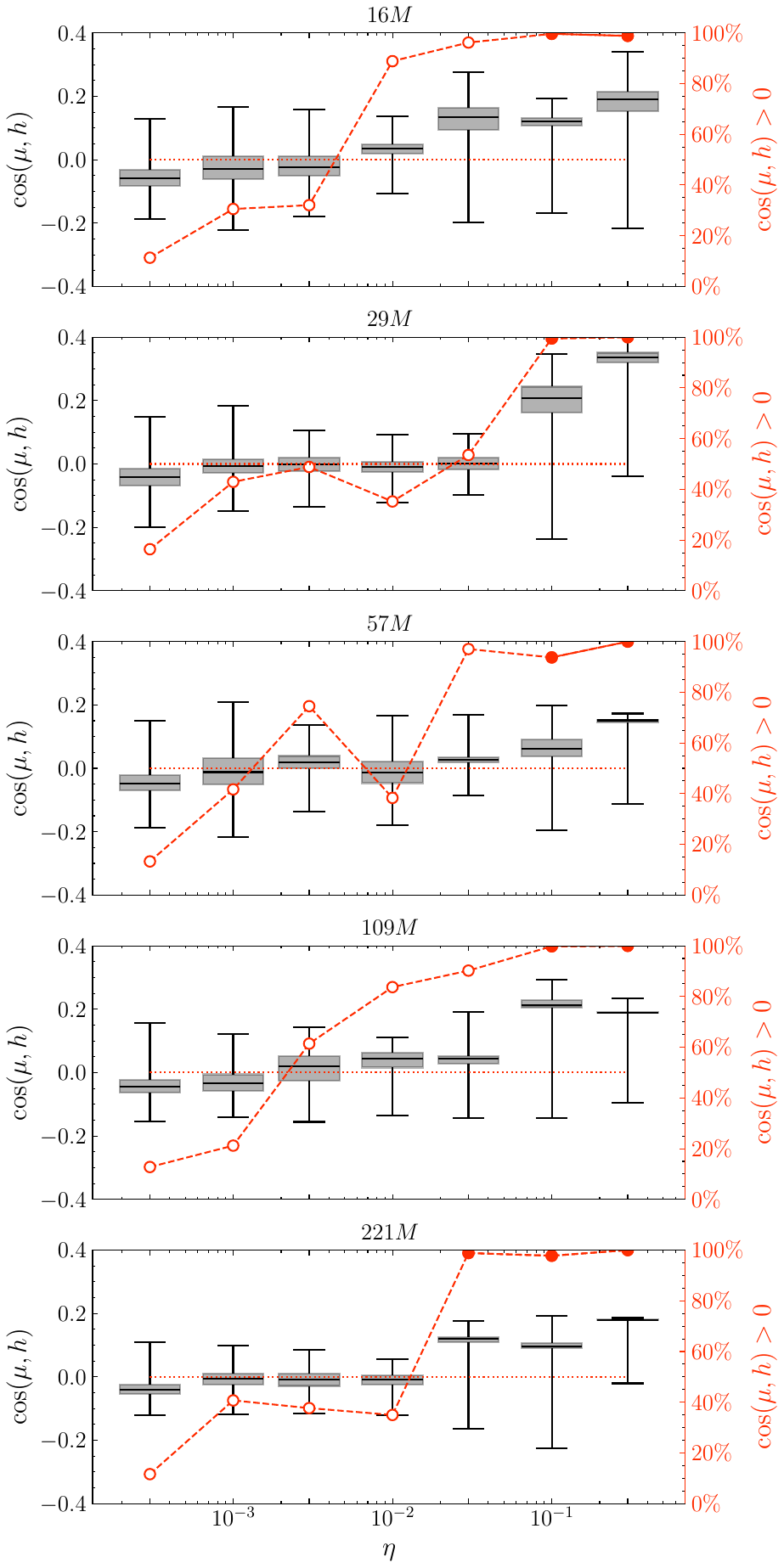}
    \caption{Results for $\ratioembeddingi$ (left) and $\cos \left( \meanembedding, \h \right)$ (right) using the baseline models without weight tying. The five rows represent the different model sizes $N$, see Eq.~\ref{eq:model_sizes}. For each combination of model size $N$ and learning rate $\eta$, the gray box represents the quartiles while the black, horizontal lines show the median and extrema. The red points display the percentage of data points that fulfill the condition given as the label of the y-axis on the right hand side. Filled and empty markers indicate divergence and convergence, respectively.}
    \label{fig:detailed_results_sample_estimates_1}    
\end{figure*}

\begin{figure*}[ht]
    \centering
    \includegraphics[scale=0.48]{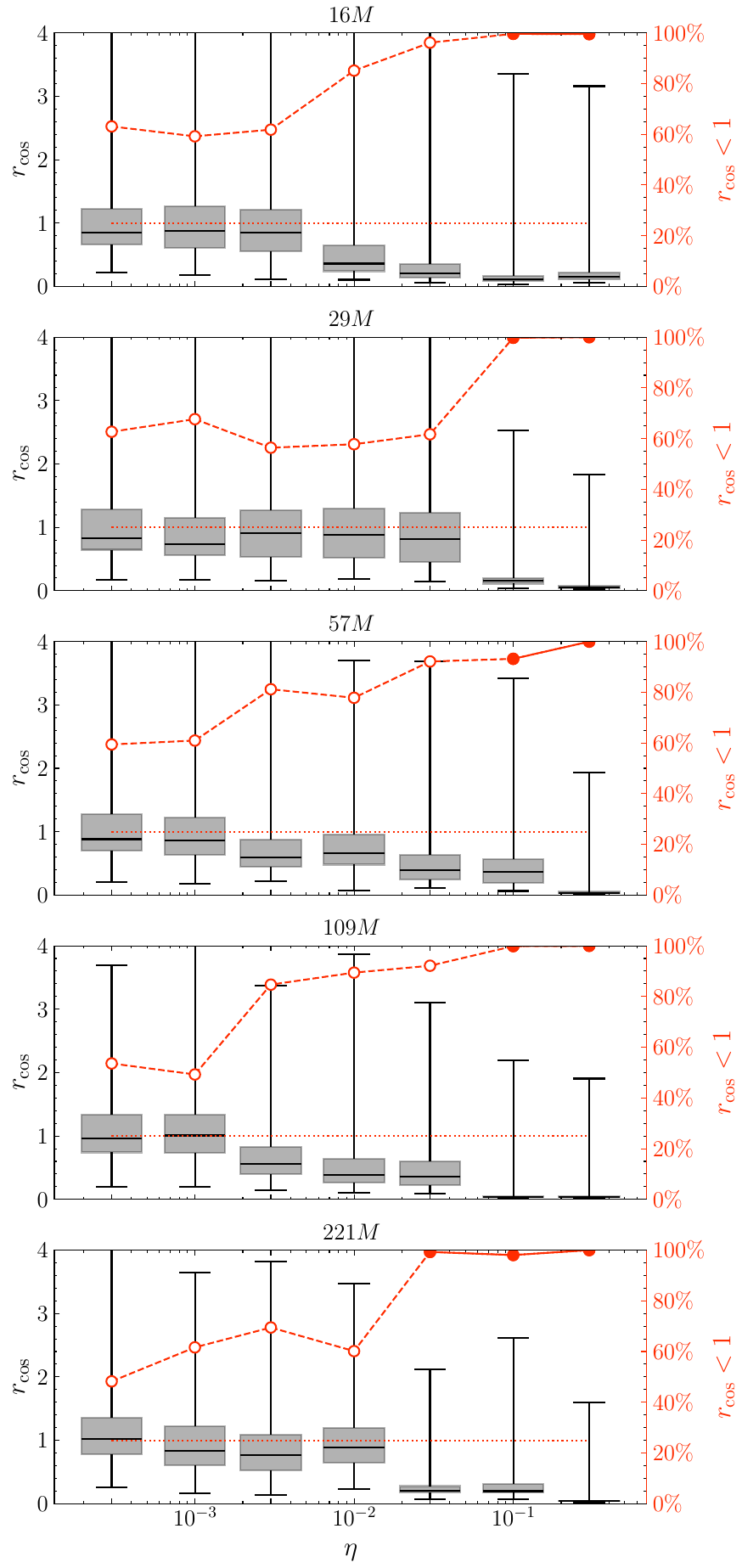}
    \hspace{1cm}
    \includegraphics[scale=0.48]{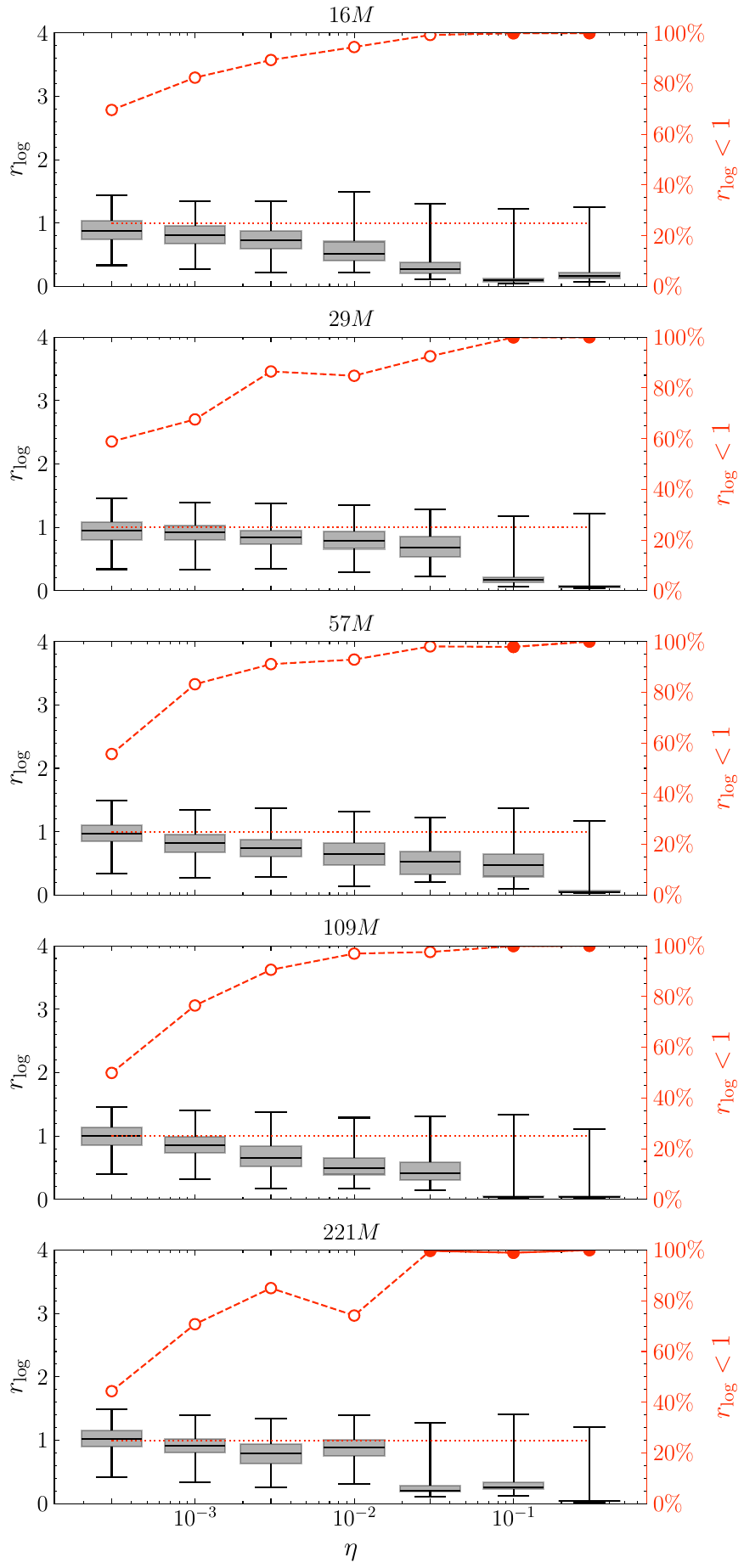}
    \caption{Results for $\ratiocosine$ (left) and $\ratiologit$ (right) using the baseline models without weight tying. The five rows represent the different model sizes $N$, see Eq.~\ref{eq:model_sizes}. For each combination of model size $N$ and learning rate $\eta$, the gray box represents the quartiles while the black, horizontal lines show the median and extrema. The red points display the percentage of data points that fulfill the condition given as the label of the y-axis on the right hand side. Filled and empty markers indicate divergence and convergence, respectively.}
    \label{fig:detailed_results_sample_estimates_2}
\end{figure*}

In Fig.~\ref{fig:wortsman_additional_2} of Sec.~\ref{sec:results}, the averages of $\ratioembeddingi$ (over the vocabulary) as well as $\cos \left( \meanembedding, \h \right)$, $\ratiocosine$, and $\ratiologit$ (over a sample of final hidden states and logits) are shown. Here, in Fig.~\ref{fig:detailed_results_sample_estimates_1} and Fig.~\ref{fig:detailed_results_sample_estimates_2}, we provide additional information on the statistical distributions of these quantities.
We find that for the vast majority of data points, the conditions are fulfilled just like for the averages. This holds true for all training runs regarding $\ratioembeddingi$, and for all diverged training runs regarding the other quantities.

\end{document}